\documentclass[10pt,twocolumn,letterpaper]{article}

\usepackage{iccv}
\usepackage{times}
\usepackage{epsfig}
\usepackage{graphicx}
\usepackage{amsmath}
\usepackage{amssymb}

\usepackage{amsfonts}
\usepackage{amsmath}
\usepackage{amsthm}
\usepackage{multirow}
\usepackage{subcaption}
\usepackage[symbol]{footmisc}

\usepackage[switch]{lineno}
\usepackage[dvipsnames]{xcolor}
\usepackage{pifont}
\usepackage[noend]{algpseudocode}
\usepackage{algorithmicx,algorithm}
\usepackage{url}

\usepackage[breaklinks=true,bookmarks=false]{hyperref}

\iccvfinalcopy 

\def\iccvPaperID{3174} 

\ificcvfinal\fi

\newtheorem{theorem}{Theorem}
\newtheorem{lemma}[theorem]{Lemma}
\newtheorem{definition}[theorem]{Definition}
\newtheorem{assumption}[theorem]{Assumption}

\begin{document}

\title{Gradient Normalization for Generative Adversarial Networks}

\author{
Yi-Lun Wu\space\space\space\space\space\space
Hong-Han Shuai\space\space\space\space\space\space
Zhi-Rui Tam\space\space\space\space\space\space
Hong-Yu Chiu\\
National Yang Ming Chiao Tung University\\
{\tt\small \{w86763777.eed08g, hhshuai, ray.eed08g, james77777778.eed07g\}@nctu.edu.tw}
}

\maketitle
\ificcvfinal\thispagestyle{empty}\fi

\begin{abstract}
   In this paper, we propose a novel normalization method called gradient normalization (GN) to tackle the training instability of Generative Adversarial Networks (GANs) caused by the sharp gradient space. Unlike existing work such as gradient penalty and spectral normalization, the proposed GN only imposes a gradient norm constraint on the discriminator function, which increases the capacity of the discriminator. Moreover, the proposed gradient normalization can be applied to different GAN architectures with little modification. Extensive experiments on four datasets show that GANs trained with gradient normalization outperform existing methods in terms of both Frechet Inception Distance and Inception Score.
\end{abstract}

\section{Introduction}

\noindent Generative Adversarial Networks (GANs)~\cite{Goodfellow2014GenerativeAN} have recently achieved great success for synthesizing new data from a given prior distribution, which facilitates a variety of applications, \eg, super resolution imaging~\cite{Ledig2017PhotoRealisticSI}, style transfer between domains~\cite{Qiao2019MirrorGANLT, Zhu2017UnpairedIT}. In the original definition, GANs consist of two networks: the \emph{generator} aims to construct realistic samples to fool the discriminator while the \emph{discriminator} learns to discriminate real samples from synthetic samples which are produced by the generator.

Although the state-of-the-art GANs generate high fidelity images that easily fool humans, the unstable training process remains a challenging problem. Therefore, a recent line of study focuses on overcoming the unstable training issue~\cite{arjovsky2017towards, Arjovsky2017WassersteinG, Gulrajani2017ImprovedTO, Salimans2016ImprovedTF, thanh2019improving}. For example, one of the reasons for the unstable GAN training is caused by the sharp gradient space of the discriminator, which causes the mode collapse in the training process of the generator~\cite{thanh2019improving}. Although L2 normalization~\cite{Kurach2019ALS} and weight clipping~\cite{Arjovsky2017WassersteinG} are simple but effective methods in stabilizing GANs, these additional constraints limit the model capacity of discriminator. As a result, the generator is inclined to fool the discriminator before learning to generate real images. Another kind of popular approaches is to formulate the discriminator as a Lipschitz continuous function bounded under a fixed Lipschitz constant $K$ by applying regularization or normalization on discriminator~\cite{Gulrajani2017ImprovedTO, Miyato2018SpectralNF, Terjek2020Adversarial, thanh2019improving}. As such, the discriminator gradient space becomes smoother without significantly sacrificing the performance of the discriminator.

Imposing a Lipschitz constraint on the discriminator can be characterized by three properties. 1) \textit{Model- or module-wise constraint.} If the constraint objective depends on full model instead of the summation of internal modules, we define such methods to be a model-wise constraint and the converse to be a module-wise constraint. For instance, the Gradient penalty (1-GP)~\cite{Gulrajani2017ImprovedTO} is a model-wise constraint, while Spectral Normalization (SN)~\cite{Miyato2018SpectralNF} is a module-wise (layer-wise) constraint, \textit{i.e.}. We argue that model-wise constraints are better since the module-wise constraints limit the layer capacities and thus reduce the multiplicative power of neural networks. 2) \textit{Sampling-based or non-sampling-based constraint.} If the approach requires sampling data from a fixed pool, such method is defined as a sampling-based constraint. For example, 1-GP is a sampling-based constraint due to the regularization, while SN is a non-sampling-based constraint since the normalization only depends on the model architecture. The non-sampling-based methods are expected to perform better than sampling-based methods since the sampling-based constraints might not be effective for the data that are not sampled before. 3) \textit{Hard or soft constraint.} If the gradient norm for the inputs of any function in the function space of the constrained discriminators is not greater than a fixed finite value, such approach is defined as a hard constraint and the converse as a soft constraint. For example, SN is a hard constraint and the fixed finite value is equal to $1$, while 1-GP is a soft constraint since the tightness of the constraint is fluctuated by the regularization and thus the upper bound is not limited. The hard constraint is expected to perform better since the consistent gradient norm can guarantee the gradient stability of unseen data.

To the best of our knowledge, none of the constraints in previous work is model-wise, non-sampling-based, and hard at the same time. In this paper, we propose a new normalization method, named gradient normalization (GN), to enforce the gradient norm bounded by 1 for the discriminator model through dividing the outputs by the gradient norm of the discriminator. Unlike SN, the proposed Lipschitz constant does not decay from the multiplicative form of neural networks since we consider the discriminator as a general function approximator and the calculated normalization term is independent of internal layers. The proposed gradient normalization enjoys the following two favorable properties. 1) The normalization simultaneously satisfies three properties including model-wise, non-sampling-based and hard constraints, and does not introduce additional hyperparameters. 2) The implementation of GN is simple and compatible with different kinds of network architectures.

\noindent The contributions of our paper are summarized as follows.
\begin{itemize}
\item In this paper, we propose a novel gradient normalization for GANs to strike a good balance between stabilizing the training process and increasing the ability of generation. To the best of our knowledge, this is the first work simultaneously satisfying the above-mentioned three properties.
\item We theoretically prove that the proposed gradient normalization is gradient norm bounded. This property helps the generator to avoid the gradient explosion or vanishing, and thus stabilizes the training process.
\item Experimental results show the proposed gradient normalization consistently outperforms the state-of-the-art methods with the same GAN architectures in terms of both Inception Score and Frechet Inception Distance. Our implementation is available at: \url{https://github.com/basiclab/GNGAN-PyTorch}.
\end{itemize}

\section{Related Work}
Previous works for addressing the issues of stabilizing GANs training mainly focus on two kinds of approaches: regularization and normalization of the discriminator. The idea behind these solutions is preventing the discriminator from generating sharp gradients during the training process. Specifically, regularization-based methods add regularization terms into the optimization process for stabilizing the training. For example, gradient penalty-based approaches~\cite{Gulrajani2017ImprovedTO, thanh2019improving, wei2018improving, Wu_2018_ECCV} compute the gradient norm from random samples as the penalty term. Lipschitz regularization~\cite{Terjek2020Adversarial} approximates maximum perturbation by power iteration. Moreover, consistency regularization~\cite{Zhang2020ConsistencyRF} regularizes the discriminator outputs of two augmented samples. Orthogonal regularization~\cite{brock2016neural} forces the square of layer weights to be an identity matrix. 

Normalization-based methods normalize each layer by different kinds of layer norms. For example, spectral normalization-based approaches~\cite{Miyato2018SpectralNF, Liu_2019_ICCV, jiang2018on} normalize layer weights by the spectral norm. Furthermore, weight normalization~\cite{xiang2017effects} normalizes layer weights by L2 norm. It is worth noting that we observed that all the normalization-based approaches are non-sampling-based and thus are more general in stabilizing the training of GANs than regularization-based methods. For example, spectral normalization divides the discriminator weight matrix by the largest singular value so that the weight matrix satisfies the Lipschitz constant constraint around 1 for all inputs. However, due to the multiplicative nature of neural networks, the final Lipschitz constant of the discriminator is decayed to a small value if the Lipschitz constant of each layer is smaller than 1. This limitation restricts the capacity of the discriminator and thus deteriorates the performance of the generative model. It is worth noting that GANs are inclined to collapse when using spectral normalization with Wasserstein loss, which is also shown in the comments from the author of spectral normalization~\cite{MiyatoComment}. We believe that the failure is caused by the imprecise dual form solution of WGAN since the discriminator function has a limited Lipschitz constant.
\section{Preliminaries}
\label{preliminaries}

\subsection{Generative Adversarial Networks}
The game between generator $G:\mathbb{R}^{d_z}\rightarrow\mathbb{R}^n$ and discriminator $D:\mathbb{R}^n\rightarrow\mathbb{R}$ can be formulated as a minimax objective~\cite{Goodfellow2014GenerativeAN}, \ie,

\begin{equation}
    \begin{aligned}
    \min_G\max_D\text{ }&\mathbb{E}_{x\sim p_r(x)}\big[log(D(x))\big]+\\
    &\mathbb{E}_{\tilde{x}\sim p_g(x)}\big[log(1-D(\tilde{x}))\big],
    \end{aligned}
    \label{GAN_objective}
\end{equation}
where $p_r(x)$ is the distribution of real data and $p_g(x)$ is the distribution defined by $p_g=G_\ast (p_z)$ ($\ast$ is the pushforward measure and $p_z$ is the $d_z$-dimensional prior distribution). In this setting, the generator $G$ is guaranteed to converge to real distribution $ p_r(x)$ if discriminator $D$ is always optimal~\cite{Goodfellow2014GenerativeAN}.
However, training GANs suffers from many difficulties including but not limited to gradient vanishing and gradient explosion. There are two major reasons causing the unstable training issues. First, optimizing the objective function~\eqref{GAN_objective} is equivalent to minimizing Jensen–Shannon divergence (JS-divergence) between $p_g(x)$ and $p_r(x)$~\cite{arjovsky2017towards, Arjovsky2017WassersteinG}. If $p_g(x)$ and $p_r(x)$ do not overlap each other, the JS-divergence would be a constant value which results in gradient vanishing. Second, finite real samples often make discriminator fall into overfitting~\cite{thanh2019improving}, which indirectly causes gradient explosion around real samples.

\subsection{Wasserstein GAN (WGAN)}
WGAN is proposed to optimize GANs by minimizing the Wasserstein-1 distance between $p_g(x)$ and $p_r(x)$, \textit{i.e.},
\begin{equation}
    \begin{aligned}
        \min_G\max_{D,L_D \le 1}\text{ }&\mathbb{E}_{x\sim p_r(x)}\big[D(x)\big]-\mathbb{E}_{\tilde{x}\sim p_g(x)}\big[D(\tilde{x})\big],
    \end{aligned}
    \label{wgan_loss}
\end{equation}
where $L_D$ is the Lipschitz constant of discriminator $D$. $L_D$ is defined as follows.
\begin{equation}
    \begin{aligned}
        L_D:=\inf\big\{&L\in\mathbb{R}:\vert D(x)-D(y) \vert\le L\Vert x - y\Vert,\\
        &\forall x,y\in\mathbb{R}^n\big\}.
    \end{aligned}
    \label{lipschitz_constant}
\end{equation}
In other words, $L_D$ is the minimum real number such that:
\begin{equation}
    \begin{aligned}
        \vert D(x)-D(y) \vert\le L_D\Vert x - y\Vert,\forall x,y\in\mathbb{R}^n.
    \end{aligned}
    \label{lipschitz_constraints}
\end{equation}
It is worth noting that the metric $\Vert\cdot\Vert$ can be any norm of vector. The discriminator in WGAN aims to approximate Wasserstein distance by maximizing the objective function~\eqref{wgan_loss} under the Lipschitz constraint~\eqref{lipschitz_constraints}. Indeed, different choices of Lipschitz constant $L_D$ do not affect the results since the Lipschitz constant of a function can be easily scaled by multiplying the function by a scaling factor. Moreover, the Wasserstein metric has been proved to be more sensible than KL metrics when the learning distributions are supported by low-dimensional manifolds~\cite{Arjovsky2017WassersteinG}. Obviously, the approximation error rate is related to the size of discriminator capacity. If a discriminator can search in a larger function space, it can approximate the Wasserstein metric more precisely, and hence makes the generator better for modeling the real distribution. Meanwhile, the Lipschitz constraint limits the steepness of value surface and therefore alleviates the overfitting of the discriminator.

However, it is still a great challenge to achieve the Lipschitz constraint on neural networks since striking a good balance between the Lipschitz constraint and network capacity is a hard task. There are many approaches proposed to achieve this constraint. Some of the approaches directly limit the Lipschitz constant for each layer, but sacrifice the function space and get limited network capacity. On the contrary, weight clipping or regularization~\cite{Arjovsky2017WassersteinG, Gulrajani2017ImprovedTO} allows networks to search in a larger function space, but loosens the constraint. In the next section, we prove that the Lipschitz constant of a layer-wise Lipschitz constraint network is upper-bounded by any of its first $k$-layer subnetworks and propose a normalization method to solve this issue.

\subsection{Notation}
\begin{definition}
    Let $f_K:\mathbb{R}^n\rightarrow\mathbb{R}$ be a $K$-layer network, which can be formulated as a function composed of a bunch of affine transformations:
    \begin{equation}
        \begin{aligned}
            f_K(x)&=\phi_K(\mathbf{W}_K\cdot(\phi_{K-1}(\cdots \mathbf{W}_1\cdot x+\mathbf{b}_1))+\mathbf{b}_K)\\
            &=\phi_K(\mathbf{W}_K\cdot f_{K-1}(x)+\mathbf{b}_K),
        \end{aligned}
    \end{equation}
    where $\mathbf{W}_K\in\mathbb{R}^{d_{K}\times d_{K-1}}$ and $\mathbf{b}_K\in\mathbb{R}^{d_K}$ are the parameters of the $K$-th layer, $d_K$ is the target dimension of the $K$-th layer and $\phi_K$ is the non-linear element-wise activation function at layer $K$. Let $f_k,\forall k\in\{1\cdots K\}$ denote the first $k$-layer subnetwork.
    \label{network_def}
\end{definition}

In order to analyze the behavior of a layer-wise constrained network, we also define layer-wise Lipschitz networks as follows.

\begin{definition}
    Let $f_K:\mathbb{R}^n\rightarrow\mathbb{R}$ be a $K$-layer network. $f_K$ is defined to be layer-wise $L$-Lipschitz constrained if $\exists L_k\le L,\forall k\in\{1\cdots K\} $ s.t. $L_k$ is the Lipschitz constant of the $k$-th layer:
    \begin{equation}
        \Vert \mathbf{W}_k\cdot x-\mathbf{W}_k\cdot y\Vert\le L_k\Vert x-y\Vert,\forall x,y\in\mathbb{R}^{d_{k-1}}.
        \label{lipschitz_layer}
    \end{equation}
    \label{layer_wise_constrained_network_def}
    \vspace{-6mm}
\end{definition}

Since the definition of the Lipschitz constraint (Inequality~\eqref{lipschitz_constraints}) is a pairwise relation which associates only two data points at a time, the sampling-based regularizations based on Inequality~\eqref{lipschitz_constraints} cannot ensure that the Lipschitz constraint is tight enough everywhere for the discriminator. Such approaches are sampling-based and soft that may cause the gradient explosion due to the non-smooth decision boundary. Consequently, a non-pairwise condition of the Lipschitz constraint is needed for helping us to associate the Lipschitz constraints with gradients. Therefore, we propose the following lemma which associates the Lipschitz constant and the gradient norm of the discriminator.

\begin{lemma}
    Let $f:\mathbb{R}^n\rightarrow\mathbb{R}$ be a continuously differentiable function and $L_f$ be the Lipschitz constant of $f$. Then the Lipschitz constraint~\eqref{lipschitz_constraints} is equivalent to
    \begin{equation}
        \begin{aligned}
            \Vert\nabla_x f(x)\Vert\le L_f,\forall x\in\mathbb{R}^n.
        \end{aligned}
        \label{grad_constraint}
    \end{equation}
    \label{lipschitz_eq_gradnorm}
    \vspace{-6mm}
\end{lemma}
\begin{proof}
See Appendix~\ref{appendix_theoretical_results} in the supplementary material.
\end{proof}
It is worth noting that the equivalence in Lemma~\ref{lipschitz_eq_gradnorm} holds if and only if the underlying function $f$ is continuous from the perspective of multivariate functions. Practically, almost all the neural networks have finite discontinuous points and therefore are continuous functions. We extend this observation and give a more useful assumption for characterizing such thorny points.

\begin{assumption}
    Let $f:\mathbb{R}^n\rightarrow\mathbb{R}$ be a continuous function which is modeled by a neural network, and all the activation functions of network f are piecewise linear. Then the function $f$ is differentiable almost surely.
    \label{network_differentiable}
\end{assumption}

This can be heuristically justified as the implementations of neural networks on a computer are subject to numerical error anyway. For complex arithmetic operations, \eg, matrix multiplication, the numerical errors are accumulated and therefore the output values are perturbed to avoid the non-differentiable points. 
\section{Gradient Normalization}
\label{gradient_normalization_section}

\setlength{\tabcolsep}{3pt}
\renewcommand{\arraystretch}{1.5}
\begin{table}[ht]
\centering
\caption{Summary of different $\zeta(x)$ used in gradient normalization~\eqref{gradient_normalization}. The derivation of the gradient norm is based on the assumption that the activation functions in the discriminator are piecewise linear. Furthermore, the sign \ding{55} means there is no potential issue of $\hat{f}(x)$ in this condition.}
\begin{tabular}{c|c|c|c}
    $\zeta(x)$ & $\vert\nabla\hat{f}(x)\vert$ & $f(x)\rightarrow\pm\infty$ & $\Vert\nabla f(x)\Vert\rightarrow 0$ \\
    \hline
    \hline
    $0$   & $1$ & \ding{55} & $\hat{f}(x)$ explodes \\
    \hline
    $1$   & $\frac{\Vert\nabla f(x)\Vert}{\Vert\nabla f(x)\Vert+1}$ &  $\hat{f}(x)$ explodes & \ding{55} \\
    \hline
    $\vert f(x)\vert$ & $\frac{\Vert\nabla f(x)\Vert}{\Vert\nabla f(x)\Vert+\vert f(x)\vert}$ & \ding{55}  & \ding{55}
    \label{gradient_normalization_summarization}
\end{tabular}
\end{table}

Lemma~\ref{lipschitz_eq_gradnorm} motivates us to design a normalization technique by directly constraining the gradient norm. We first show that the Lipschitz constant of a layer-wise $1$-Lipschitz constraint, \eg, SN-GAN, may significantly decrease when the number of layers increases, which inspires the concept of the proposed gradient normalization. In the following, we assume that the activation functions are 1-Lipschitz functions\footnote{Activation functions commonly-used, \eg, ReLU, Leaky ReLU, SoftPlus, Tanh, Sigmoid, ArcTan, Softsign are 1-Lipschitz functions~\cite{LipschitzRegularity18}.} and prove that the Lipschitz constant of a deeper network is bounded by its shallow network.
\begin{theorem}
    \label{lipschitz_decrease}
    Let $f_K:\mathbb{R}^n\rightarrow\mathbb{R}$ be a layer-wise 1-Lipschitz constrained $K$-layer network. The Lipschitz constant of the first $k$-layer network $L_{f_k}$ is upper-bounded by $L_{f_{k-1}}$, \ie,
    \begin{equation}
        L_{f_k}\le L_{f_{k-1}},\forall k\in \{2\cdots K\}.
        \label{lipschitz_inequality}
    \end{equation}
\end{theorem}
\begin{proof}
See Appendix~\ref{appendix_theoretical_results} in the supplementary material.
\end{proof}

In Theorem \ref{lipschitz_decrease}, the equality \eqref{lipschitz_inequality} holds if and only if $\exists x,y\in\mathbb{R}^n$ $s.t.$
\begin{equation}
    \begin{aligned}
        \frac{\Vert f_k(x)-f_k(y)\Vert}{\Vert x-y\Vert}&=\frac{\Vert f_{k-1}(x)-f_{k-1}(y)\Vert}{\Vert x-y\Vert}=L_{f_{k-1}}.
    \end{aligned}
\end{equation}
It is difficult to ensure that the equation holds, especially when the network is optimized by stochastic-gradient-based methods. If the equality does not hold at the $k$-th layer, the Lipschitz constant of the first $k$-layer subnetwork is less than the Lipschitz constant of the first $(k-1)$-layer subnetwork. By applying this rule iteratively, we have
\begin{equation}
    L_{f_k}<L_{f_{k-1}}<\cdots<L_{f_1}\le 1.
    \label{lipschitz_absolutly_decrease}
\end{equation}
Therefore, the Lipschitz constant may be drastically reduced when the number of layers increases. It is worth noting that this is a potential reason which makes SN-GAN fail to integrate with Wasserstein distance when the gradient-based regularization is not applied. On the other hand, the Lipschitz constrained networks do not need to be layer-wise Lipschitz constrained, i.e., instead of a module-wise constraint, it is possible to build a network which is constrained by a model-wise characteristic of a network .

Inspired by this observation and Lemma~\ref{lipschitz_eq_gradnorm}, we propose a normalization method which strictly limits the gradient norm but also maintains a high discriminator capacity. Specifically, as shown in Lemma \ref{lipschitz_eq_gradnorm}, the Lipschitz constant is associated with the gradient norm. We then propose a new solution, Gradient Normalization (GN), to make the network search in a function space induced by constraint \eqref{grad_constraint}. Let $f:\mathbb{R}^n\rightarrow\mathbb{R}$ be a continuously differentiable function, the proposed GN normalizes the norm of the gradient $\Vert \nabla_x f(x)\Vert$ and bounds $f(x)$ simultaneously:
\begin{equation}
    \hat{f}(x)=\frac{f(x)}{\Vert\nabla_x f(x)\Vert+\zeta(x)},
    \label{gradient_normalization}
\end{equation}
where $\zeta(x):\mathbb{R}^n\rightarrow\mathbb{R}$ is a universal term which can be associated with $f(x)$ or a constant to avoid the situation that $|\hat{f}(x)|$ approximates infinity or $\Vert\nabla_x \hat{f}(x)\Vert$ approximates $0$. Here, we propose to set $\zeta(x)$ as $\vert f(x)\vert$ and prove that this gradient normalization is gradient norm bounded. After that, the variants of $\zeta(x)$ are discussed for explanations.

\begin{theorem}
    Let $f:\mathbb{R}^n\rightarrow\mathbb{R}$ be a continuous function which is modeled by a neural network, and all the activation functions of network $f$ are piecewise linear. The normalized function $\hat{f}(x)=f(x)/\big(\Vert\nabla_x f(x)\Vert+\vert f(x)\vert\big)$ is gradient norm bounded, \ie,
    \begin{equation}
        \Vert\nabla_x\hat{f}(x)\Vert=\Bigg\vert\frac{\Vert\nabla f\Vert}{\Vert\nabla f\Vert+\vert f\vert}\Bigg\vert^2\le 1.
        \label{normalized_gradient_norm}
    \end{equation}
    \label{bounded_gradient_normalization_theorem}
\end{theorem}
\begin{proof}
See Appendix~\ref{appendix_theoretical_results} in the supplementary material.
\end{proof}

The potential issues of two basic approaches for $\zeta(x)$, \ie, $\zeta(x)=0$ and $\zeta(x)=1$, are summarized in Table~\ref{gradient_normalization_summarization}. Considering the overfitting of a gradient normalized discriminator, the discriminator gives out confident predictions for both real and fake samples. From Eq.\eqref{gradient_normalization}, the confident predictions are produced by $f(x)\rightarrow\pm\infty$ and $\Vert\nabla_x f(x)\Vert\rightarrow 0$, which are also the conditions shown in Table~\ref{gradient_normalization_summarization}. Since the function norm $\vert f(x)\vert$ is not directly related to the gradient norm $\Vert\nabla_x f(x)\Vert$, the confident predictions may make normalized gradient norm $\Vert \nabla_x\hat{f}(x)\Vert$ and normalized function value $\vert\hat{f}(x)\vert$ explode\footnote{The empirical results will be discussed later in Section~\ref{ablation_study}}. To deal with this problem, therefore, we propose the formulation which sets $\zeta(x)$ to $\vert f(x) \vert$. In this setting, when the discriminator is saturated due to overfitting, the normalized gradient norm~\eqref{normalized_gradient_norm} is close to $0$. This self-control mechanism prevents the generator from getting an exploded gradient and consequently stabilizes the training process of GANs. The pseudocode of the proposed GN is presented in Algorithm~\ref{gn_gan_training}.

\begin{algorithm}
    \caption{Gradient Normalized GAN (GN-GAN)}
    \label{gn_gan_training}
    {\bf Input:} generator and discriminator parameter $\theta_G$, $\theta_D$, learning rate $\alpha_G$ and $\alpha_D$, batch size $M$, the number of discriminator updates per generator updates $N_{dis}$, total iterations $N$
    \begin{algorithmic}[1]
    \State $\hat{D}:=D(x)/(\Vert\nabla_x D(x)\Vert+\vert D(x)\vert)$
    \For{$t = 1$ to $N$}
        \For{$t_{dis} = 1$ to $N_{dis}$}
            \For{$i = 1$ to $M$}
                \State $z\sim p_z,x\sim p_r(x)$
                \State $\mathcal{L}^i_D\leftarrow\hat{D}(G(z))-\hat{D}(x)$
            \EndFor
            \State $\theta_D\leftarrow Adam\big(\frac{1}{2M}\sum_{i=1}^{M}\mathcal{L}^i_D,\alpha_D,\theta_D\big)$
        \EndFor
        \State Sample a batch of latent variable $\{z_i\}_{i=1}^{2M}\sim p_z(z)$
        \State $\theta_G\leftarrow Adam\bigg(\frac{1}{2M}\sum_{i=1}^{2M}\Big(-\hat{D}\big(G(z_i)\big)\Big),\alpha_G,\theta_G\bigg)$
    \EndFor
    \end{algorithmic}
\end{algorithm}

\noindent\textbf{Gradient Analysis of Gradient Normalization.}
The gradient of $\hat{f}(x)$ with respect to $W_k$ is derived as follows. Please note that $\zeta(x)$ is set as $\vert f(x)\vert$ and function arguments are ignored here for simplicity.
\begin{subequations}
    \label{gradient_of_gradient_normalization}
    \begin{align}
        \label{gradient_of_gradient_normalization:1}
        \frac{\partial\hat f}{\partial\mathbf{W}_k}=&\frac{\partial\hat f}{\partial f}\frac{\partial f}{\partial\mathbf{W}_k}+\frac{\partial\hat f}{\partial\Vert\nabla_x f\Vert}\frac{\partial\Vert\nabla_x f\Vert}{\partial\mathbf{W}_k}\\
        \label{gradient_of_gradient_normalization:2}
        \begin{split}
        =&\frac{\Vert\nabla_x f\Vert}{\big(\Vert\nabla_x f\Vert+\vert f\vert\big)^2}\frac{\partial f}{\partial\mathbf{W}_k}-\\
        &\frac{f}{\big(\Vert\nabla_x f\Vert+\vert f\vert\big)^2}\frac{\partial \Vert\nabla_x f\Vert}{\partial\mathbf{W}_k}
        \end{split}\\
        \label{gradient_of_gradient_normalization:3}
        =&\frac{1}{\big(\Vert\nabla_x f\Vert+\vert f\vert\big)^2}\Bigg(\Vert\nabla_x f\Vert\frac{\partial f}{\partial\mathbf{W}_k}-f\frac{\partial\Vert\nabla_x f\Vert}{\partial\mathbf{W}_k}\Bigg).
    \end{align}
\end{subequations}
Interestingly, according to Eq.\eqref{gradient_of_gradient_normalization:3}, the gradient normalization is a special form of an adaptive gradient regularization. More specifically, in Eq.\eqref{gradient_of_gradient_normalization:1}, the first term is the gradient of GAN objective which improves the discriminating power, while the second term is the regularization which penalizes the gradient norm of $f$ with the adaptive regularization coefficient. Compared with 0-GP~\cite{thanh2019improving} and 1-GP~\cite{Gulrajani2017ImprovedTO}, the gradient penalty in GN is more flexible and can automatically negotiate with GAN loss. Consequently, this self-balancing mechanism forces the gradient norm of GN to be bounded.

\noindent\textbf{Comparisons of Different Approaches.} Table~\ref{summarization} summarizes these three properties of several well-known methods. For the regularization-based methods~\cite{brock2016neural,Gulrajani2017ImprovedTO,thanh2019improving,Zhang2020ConsistencyRF}, the constraints are always soft due to the trade-off between the regularization and the objective of GANs. Furthermore, the non-sampling-based approaches~\cite{brock2016neural,Miyato2018SpectralNF} consider the network as a function composed of multiple layers and impose the constraints on individual layers to achieve the Lipschitz constraint on the full model. By Theorem~\ref{lipschitz_decrease}, these layer-wise constraints potentially reduce the capacity of the discriminator and therefore sacrifice the generation quality. In contrast, the gradient normalization does not depend on the specific subset of data and is applicable on the full model. Accordingly, by Theorem~\ref{bounded_gradient_normalization_theorem}, such normalization is a model-wise, non-sampling-based and hard constraint.

\setlength{\tabcolsep}{2pt}
\renewcommand{\arraystretch}{1.0}
\begin{table}[t]
\centering
\caption{Summary of different regularization and normalization techniques.}
\begin{tabular}{c|c|c|c}
    \textbf{Method} & \textbf{Model-wise} & \textbf{Non-sampling-based} & \textbf{Hard} \\
    \hline
    Orthogonal~\cite{brock2016neural}              &            & \checkmark &             \\
    \hline
    1-GP~\cite{Gulrajani2017ImprovedTO}                           & \checkmark &                &             \\
    \hline
    0-GP~\cite{thanh2019improving}                                & \checkmark &                &             \\
    \hline
    CR~\cite{Zhang2020ConsistencyRF} & \checkmark &                &           \\
    \hline
    SN~\cite{Miyato2018SpectralNF}       &            & \checkmark & \checkmark \\
    \hline
    (our) GN                                  & \checkmark & \checkmark & \checkmark
    \label{summarization}
\end{tabular}
\end{table}
\section{Experiments}

To evaluate gradient normalization, we first conduct the experiments of unconditional and conditional image generation on two standard datasets, namely CIFAR-10 dataset~\cite{Coates2011AnAO} and STL-10 dataset~\cite{Torralba200880MT}. CIFAR-10 dataset contains 60k images of size $(32\times32\times3)$ which was partitioned into 50k training instances and 10K testing instances. STL-10 dataset is designed for developing unsupervised feature learning, containing 5k training images, 8k testing images and 100k unlabeled set of size $(48\times48\times3)$. Moreover, we also test the proposed method on two datasets with a higher resolution including CelebA-HQ~\cite{karras2017progressive} and LSUN~\cite{yu15lsun} Church Outdoor. CelebA-HQ contains 30k human faces of size $(256\times256\times3)$ and LSUN Church Outdoor is a subset of the LSUN dataset containing 126k church outdoor scenes of size $(256\times256\times3)$.

\renewcommand{\arraystretch}{1.18}
\begin{table*}[hbt!]
\centering
\caption{Inception Score and FID with unconditional image generation on CIFAR-10 and STL-10. We report the average and standard deviation of the results trained with 5 different random seeds. Note that ``-'' denotes that result is not reported by the original paper. Moreover, $\dagger$ represents that the original paper does not provide an evaluation on STL-10, so we provide a implementation for reference. $\ddagger$ denotes that we provide a re-implementation result for reliable comparison.}
\label{unconditional_image_generation_result}
\begin{tabular}{lccccccc}
    \hline
    \multirow{2}{*}{Method} & \multicolumn{3}{c}{\textbf{CIFAR-10}} & \multicolumn{2}{c}{\textbf{STL-10}} \\
                         & Inception Score$\uparrow$ & FID(train)$\downarrow$ & FID(test)$\downarrow$ & Inception Score$\uparrow$ & FID(50k) $\downarrow$ & FID(10k) $\downarrow$\\
    \hline
    Real data & 11.24$\pm$.12 & 0 & 7.80 & 26.08$\pm$.26 & 0 & 0\\
    \hline
    \multicolumn{6}{l}{\textit{\textbf{Standard CNN}}} \\
    SN-GAN~\cite{Miyato2018SpectralNF} & 7.58$\pm$.12 & - & 25.50 & 8.79$\pm$.14 & - & 43.20 \\
    SN-GAN$^\ddagger$ & 7.86$\pm$.09 & 18.52 & 22.67 & 8.87$\pm$.09 & 32.90 & 35.10 & \\
    SN-GAN-CR~\cite{Zhang2020ConsistencyRF} & 7.93 & - & 18.72 & 8.69$\pm$.08$^\dagger$ & 32.11$^\dagger$ & 34.14$^\dagger$ \\
    (our) GN-GAN & 7.71$\pm$.14 & 19.31$\pm$.76 & 23.52$\pm$.80 & 9.00$\pm$.15 & 30.18$\pm$.82 & 32.41$\pm$.73 \\
    (our) GN-GAN-CR & 8.04$\pm$.19 & 18.59$\pm$1.5 & 22.89$\pm$1.5 & 9.00$\pm$.14 & 27.61$\pm$.69 & 29.53$\pm$.62 \\
    \hline
    \multicolumn{6}{l}{\textit{\textbf{ResNet}}} \\
    WGAN-GP~\cite{Gulrajani2017ImprovedTO} & 7.86$\pm$.08 & - & - & - & - & - \\
    SN-GAN & 8.22$\pm$.05 & - & 21.70$\pm.21$ & 9.10$\pm$.04 & - & 40.10$\pm$.50 \\
    SN-GAN$^\ddagger$ & 8.48$\pm$.11 & 12.35 & 16.59 & 9.18$\pm$.10 & 29.16 & 31.85 \\
    SN-GAN-CR & 8.40 & - & 14.56 & 9.38$\pm$.07$^\dagger$ & 25.78$^\dagger$ & 28.4$^\dagger$ \\
    (our) GN-GAN & 8.49$\pm$.11 & 11.13$\pm$.18 & 15.33$\pm$.16 & 9.60$\pm$.14 & 26.14$\pm$.7 & 28.12$\pm$0.61 \\
    (our) GN-GAN-CR & \textbf{8.72$\pm$.11} & \textbf{9.55$\pm$.47} & \textbf{13.71$\pm$.40} & \textbf{9.74$\pm$.15} & \textbf{23.62$\pm$.89} & \textbf{25.80$\pm$0.59} \\
    \hline
    \multicolumn{6}{l}{\textit{\textbf{Neural Architecture Search}}} \\
    AutoGAN~\cite{Gong_2019_ICCV} & 8.55$\pm$.10 & 12.42 & - & 9.16$\pm$.12 & 31.01 & - \\
    E$^2$GAN~\cite{Tian_2020_ECCV} & 8.51$\pm$.13 & 11.26 & - & 9.51$\pm$.09 & 25.53 & - \\
    \hline
\end{tabular}
\end{table*}

Two popular evaluation metrics for generative models, \ie, Inception Score~\cite{Salimans2016ImprovedTF} and Frechet Inception Distance (FID)~\cite{Heusel2017GANsTB}, are used to quantitatively evaluate the proposed method. For a fair comparison, all evaluations are computed by the official implementation of the Inception Score and FID. In order to compare with previous works which did not carefully follow the standard evaluation protocol, our evaluation is designed according to several different configurations for calculating FID~\cite{Gong_2019_ICCV, Kurach2019ALS}. Furthermore, we record the best model checkpoint in terms of FID throughout the training and report the averaged results. For more details of evaluation, see Appendix~\ref{appendix_evaluation_details} in the supplementary material.

\subsection{Unconditional Image Generation}
\label{unconditional_image_generation_section}
Table~\ref{unconditional_image_generation_result} compares the proposed GN-GAN with several state-of-the art models, \eg, SN-GAN~\cite{Miyato2018SpectralNF}, WGAN-GP~\cite{Gulrajani2017ImprovedTO} and CR-GAN~\cite{Zhang2020ConsistencyRF}. The results manifest that the proposed GN-GAN outperforms existing normalization methods in terms of Inception Score and FID. We also combine GN-GAN with consistency regularization (CR)~\cite{Zhang2020ConsistencyRF}, named GN-GAN-CR, by directly replacing spectral normalization with GN. The result indicates that GN-GAN-CR can further improve SN-GAN-CR in both Inception Score and FID by simply replacing the normalization method. It is worth noting that the Wasserstein loss degenerates into hinge loss in the proposed GN-GAN since GN makes the output of network range within $[-1,1]$. Therefore, we only test our model with two loss functions: hinge loss and non-saturating loss (NS)~\cite{Goodfellow2014GenerativeAN}. We derive the best performance by using hinge loss with ResNet and NS loss with Standard CNN. The architectures and most of optimization parameters in our experiments are the same as~\cite{Miyato2018SpectralNF}. We stop training after 200k generator update steps for all the experiments shown in Table~\ref{unconditional_image_generation_result}. The learning rate of the discriminator is slightly increased from $2\times10^{-4}$ to $4\times10^{-4}$ for ResNet architecture. In GN-GAN-CR, we test regularization weight $\lambda=0.1, 1, 5, 10$ on CIFAR-10 with ResNet and find that $\lambda=5$ performs better than the others.

\subsection{Conditional Image Generation}
We further conduct the experiments on the CIFAR-10 dataset with the same architecture proposed by BigGAN~\cite{Brock2019LargeSG}. Similarly, SN in the discriminator is replaced by GN.\footnote{Following BigGAN, SN is still used in the generator.} Table~\ref{conditional_image_generation} shows that GN can further improve BigGAN by 31.8\% in terms of FID. It is worth noting that we increase the discriminator learning rate from $1\times10^{-4}$ to $2\times10^{-4}$, which is two times greater than that of the generator during training. This modification is motivated by the self-control mechanism of GN, which makes the outputs of the discriminator saturate thus needing more steps to give confident predictions.

\renewcommand{\arraystretch}{1.15}
\begin{table}[bt!]
    \centering
    \caption{Inception Score and FID with conditional image generation on CIFAR-10.}
    \begin{tabular}{lcc}
        \hline
        Method & Inception Score$\uparrow$ & FID(test)$\downarrow$ \\
        \hline
        BigGAN & 9.22 & 14.73 \\
        BigGAN-CR~\cite{Zhang2020ConsistencyRF} & - & 11.48 \\
        (our) GN-BigGAN & 9.22$\pm$.13 & \textbf{10.05$\pm$.23} \\
        \hline
    \end{tabular}
    \label{conditional_image_generation}
\end{table}

\subsection{Unconditional Large Scale Image Generation}
\begin{figure}[b]
    \centering
    \includegraphics[scale=0.46]{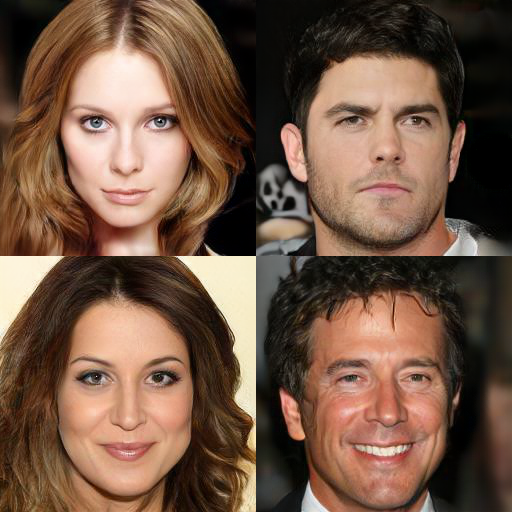}
    \caption{Generated samples on CelebA-HQ 256$\times$256. FID=7.67.}
    \label{large_scale_image_celebhq}
\end{figure}

\begin{figure}[b]
    \centering
    \includegraphics[scale=0.46]{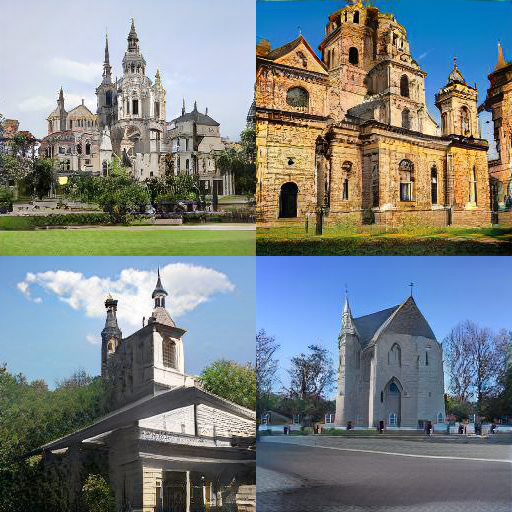}
    \caption{Generated smaples on LSUN Church Outdoor 256$\times$256. FID=5.405.}
    \label{large_scale_image_lsun_church}
\end{figure}

To show that the proposed Gradient Normalization is able to generate high-resolution images, we leverage the architecture proposed by SN-GAN for generating 256$\times$256 images on CelebA-HQ and LSUN Church Outdoor. Similarly, SN is replaced by GN in the experiments. In Figures~\ref{large_scale_image_celebhq} and \ref{large_scale_image_lsun_church}, generated results show the competitive quality. Due to the space constraint, more samples and quantitative results are provided in Appendix~\ref{appendix_experiments_details} of the supplementary materials.

\begin{figure*}[tbh]
    \begin{subfigure}{0.33\textwidth}
        \centering
        \includegraphics[scale=0.4]{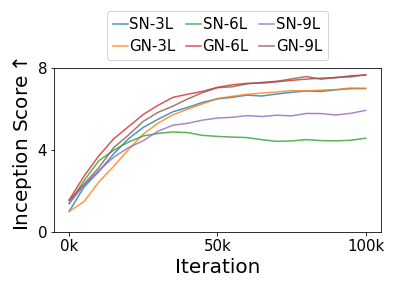}
        \caption{}
        \label{layer_wise_vs_global:vis_IS_vs_CNN}
    \end{subfigure}%
    \begin{subfigure}{0.33\textwidth}
        \centering
        \includegraphics[scale=0.4]{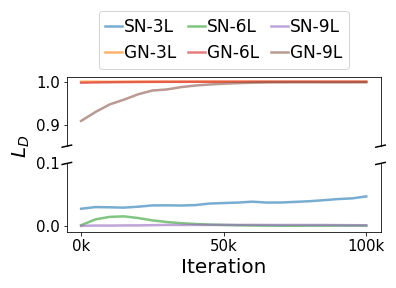}
        \caption{}
        \label{layer_wise_vs_global:vis_lipschitz_vs_depth}
    \end{subfigure}%
    \begin{subfigure}{0.33\textwidth}
        \centering
        \includegraphics[scale=0.4]{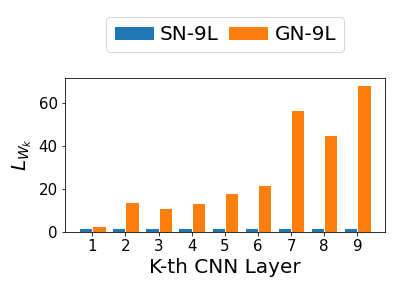}
        \caption{}
        \label{layer_wise_vs_global:vis_lipschitz_vs_layer}
    \end{subfigure}%
    \label{layer_wise_vs_global}
    \caption{Analysis of Theorem~\ref{lipschitz_decrease} on CIFAR-10 dataset. Best viewed in color.}
\end{figure*}

\subsection{Experimental Analysis for Theorem~\ref{lipschitz_decrease}}
According to Theorem~\ref{lipschitz_decrease}, the Lipschitz constant decreases as the number of layers increases. To explain the results clear, we first introduce Corollary 1 in WGAN-GP~\cite{Gulrajani2017ImprovedTO}, which states that if the GAN is trained with Wasserstein loss, there exists at least one optimal discriminator $D^*$ satisfying $\Vert\nabla_xD^*(x)\Vert=1$ almost everywhere under the support of $p_r$ and $p_g$. Accordingly, by Lemma~\ref{lipschitz_eq_gradnorm}, the Lipschitz constant for such optimal discriminator is equal to $1$ almost surely. We therefore design the experiments to test SN and GN with the Wasserstein loss. Figures~\ref{layer_wise_vs_global:vis_IS_vs_CNN} and~\ref{layer_wise_vs_global:vis_lipschitz_vs_depth} respectively show the Inception Scores and the Lipschitz constants of discriminators with regards to the training iterations for different approaches on CIFAR-10 dataset, where $n$L means both generator and discriminator are modeled by $n$ convolutional layers and $L_D$ is approximated by the maximum gradient norm of 50k sampling data from each of $p_d$ and $p_g$. However, Figures~\ref{layer_wise_vs_global:vis_IS_vs_CNN} and~\ref{layer_wise_vs_global:vis_lipschitz_vs_depth} show that the generators of SN-GANs do not lead to a high IS and the Lipschitz constants of discriminators are much smaller than $1$. Moreover, all the discriminators of SN-GANs cannot well approximate Wasserstein distance under the Lipschitz constraint $L_D\le 1$ even if the above corollary guarantees the existence of optimal discriminator. We believe the reason is that the SN over-restricts the Lipschitz constants of discriminators, and hence discriminators cannot increase the Lipschitz constants of themselves to approach $D^*$. This situation becomes more worse when the number of layers is increased (from $3$L to $9$L). Conversely, the magnitude of Lipschitz constants of GN-GANs is invariant to the depth of model and leads to a better performance, which indeed matches the Theorem~\ref{lipschitz_decrease}.

We further investigate the Lipschitz constant of internal layers.
Figure~\ref{layer_wise_vs_global:vis_lipschitz_vs_layer} shows the Lipschitz constants of GN-9L and SN-9L in the layer level on CIFAR-10 dataset. It is worth noting that GN can achieve 1-Lipschitz constraint without constraining internal layers. Thus, the Lipschitz constant of each layer in GN-9L is unlimited and is more flexible than module-wise approach, \eg, SN-9L. The multiplicative power of discriminators, therefore, is not limited, which is the potential superiority that makes GN converge well with Wasserstein loss.

\subsection{Ablation Study}
\label{ablation_study}
\begin{figure}[t]
    \centering
    \includegraphics[scale=0.20]{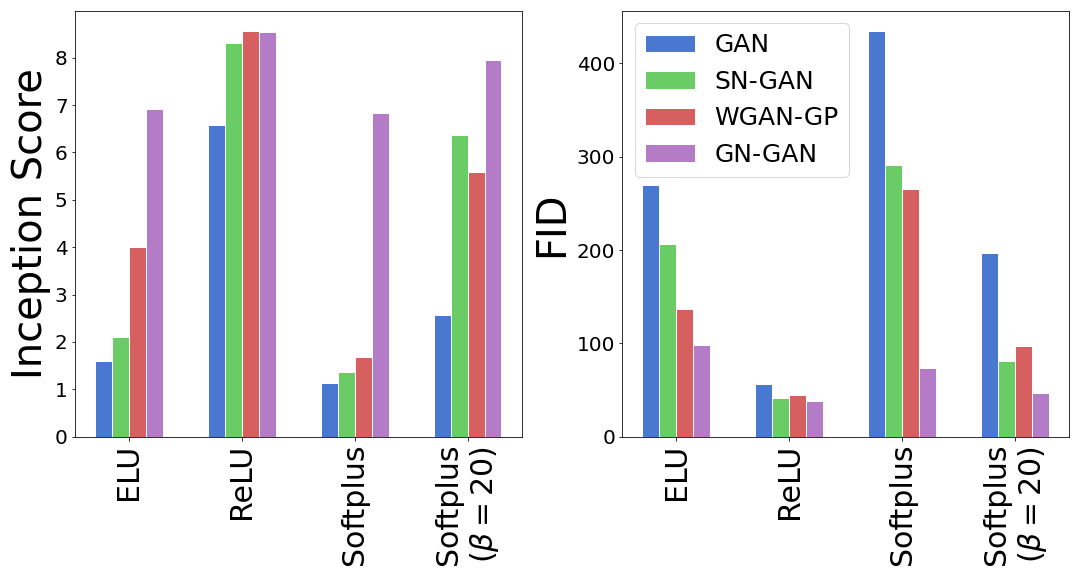}
    \caption{Comparison of activation functions including ELU~\cite{clevert2015fast}, ReLU~\cite{nair2010rectified} and Softplus~\cite{glorot2011deep} on STL-10.}
    \label{ablation_activation_stl10}
\end{figure}

\noindent\textbf{Activation Function.} Theorem~\ref{bounded_gradient_normalization_theorem} only provides the upper bound of gradient norm with the assumption that the activation functions of the discriminator are piecewise linear. However, we hypothesize that GN works for most activation functions. Therefore, we reproduce the Standard CNN training reported by SN-GAN~\cite{Miyato2018SpectralNF}, WGAN-GP~\cite{Gulrajani2017ImprovedTO} and Vanilla GAN~\cite{Goodfellow2014GenerativeAN} with different activation functions. Figure~\ref{ablation_activation_stl10} shows the performance of GAN, WGAN-GP, SN-GAN and GN-GAN with different activation functions in terms of IS and FID, which indicates that GN achieves the best scores on the ELU and ReLU, while gets competitive scores on Softplus ($\beta=20$). It is worth noting that Softplus becomes similar to ReLU if the $\beta$ increases. As such, GN performs better on Softplus($\beta=20$) than Softplus.

\begin{figure}[t]
    \centering
    \includegraphics[scale=0.20]{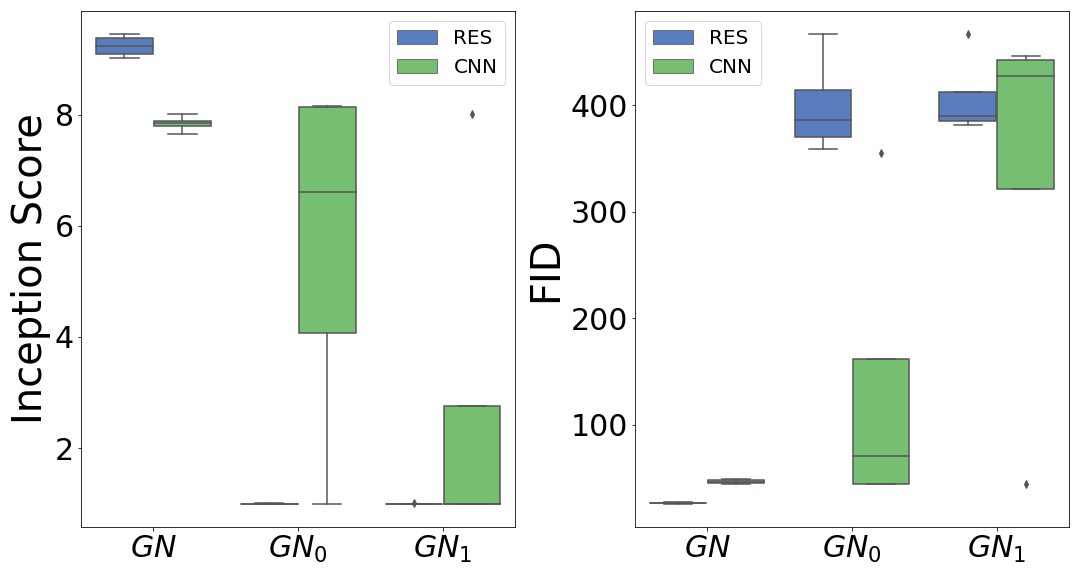}
    \caption{Comparison of variants of gradient normalization~\eqref{gradient_normalization} on STL-10. The experiments include $\zeta(x)=\vert f(x)\vert$ ($GN$), $\zeta(x)=1$ ($GN_1$) and $\zeta(x)=0$ ($GN_0$).}
    \label{ablation_formulation_stl10}
\end{figure}

\noindent\textbf{Variants of Gradient Normalization.} As discussed in Section~\ref{gradient_normalization_section}, we set $\zeta(x)$ to $\vert f(x)\vert$ in all experiments for stability issues. Here, we compare three variants of gradient normalization, $\zeta(x)=\vert f(x)\vert$ ($GN$), $\zeta(x)=1$ ($GN_1$) and $\zeta(x)=0$ ($GN_0$), by applying them for unconditional image generation. Figure~\ref{ablation_formulation_stl10} shows the results of three normalization approaches across different model architectures and datasets with the total iteration as 50k. The hyperparameter used for different $\zeta(x)$ is the same as~\ref{unconditional_image_generation_section}. Moreover, we repeat the training process 4 times with different random seeds and report the average performance of the last checkpoints. The results manifest that the variance of Inception Score and FID for $GN_0$ is large for different architectures and datasets. Even if we add a constant to the denominator of~\eqref{gradient_normalization} by setting $\zeta(x)$ to a non-zero constant $1$, the training results of $GN_1$ are still inferior to the proposed $GN$. These results match our discussion in Section~\ref{gradient_normalization_section}.

\section{Conclusion and Future Work}

In this paper, we propose a novel gradient normalization method for stabilizing the training of GANs, which can facilitate a variety of applications. The proposed GN is simple to implement and theoretically proven to generate a bounded gradient norm. Also, we apply GN to several different architectures on different datasets and most of them achieve state-of-the-art results. In the future, we plan to replace the gradient norm of the denominator term with a quasi formulation for further reducing the computation. Another interesting direction is to apply GN to other GAN-related tasks such as style transfer, super-resolution, and video generation.

\section*{ACKNOWLEDGMENT}
This work is supported in part by the Ministry of Science and Technology (MOST) of Taiwan under the grants MOST-109-2221-E-009-114-MY3 and MOST-110-2218-E-A49-018. This work was also supported by the Higher Education Sprout Project of the National Yang Ming Chiao Tung University and Ministry of Education (MOE), Taiwan. We are grateful to the National Center for High-performance Computing for computer time and facilities.

\appendix
\section{Theoretical Results}
\label{appendix_theoretical_results}

We first define the Lipschitz constant $L_f$ again for a better readability. Let $f:\mathbb{R}^n\rightarrow\mathbb{R}$ be a mapping function. Then, $L_f$ is the minimum real number such that:
\begin{equation}
    \begin{aligned}
        \vert f(x)-f(y) \vert\le L_f\Vert x - y\Vert,\forall x,y\in\mathbb{R}^n.
    \end{aligned}
    \label{lipschitz_constraints_appendix}
\end{equation}

\noindent\textbf{Lemma 3.} \textit{
Let $f:\mathbb{R}^n\rightarrow\mathbb{R}$ be a continuously differentiable function and $L_f$ be the Lipschitz constant of $f$. Then the Lipschitz constraint} \eqref{lipschitz_constraints_appendix} is equivalent to
\begin{equation}
    \begin{aligned}
        \Vert\nabla_x f(x)\Vert\le L_f,\forall x\in\mathbb{R}^n.
    \end{aligned}
    \label{grad_constraint_appendix}
\end{equation}

\begin{proof}
We first prove the sufficient condition.\newline
    $(\Rightarrow)$ From the definition of Lipschitz constraint~\eqref{lipschitz_constraints_appendix}, we know
    \begin{equation}
        \vert f(x)-f(y)\vert\le L_f\Vert x-y\Vert.
    \end{equation}
    Now, we consider the norm of directional derivative at $x$ along with the direction of $(y-x)$:
    \begin{equation}
        \langle\nabla f(x),\frac{y-x}{\Vert y-x\Vert}\rangle=\lim_{y\rightarrow x}\frac{\vert f(y)-f(x)\vert}{\Vert x-y\Vert}\le L_{f},
    \end{equation}
    where $\langle\cdot,\cdot\rangle$ is the inner product.
    Since the norm of gradient is the maximum norm of directional derivative, then
    \begin{equation}
        \Vert\nabla f(x)\Vert\le L_{f}.
    \end{equation}
    We then prove the necessary condition.\newline
    $(\Leftarrow)$ By the assumption, $f$ is continuous and differentiable. Therefore, the conditions of \emph{Gradient theorem} are satisfied, and thus we can only consider the line integral along the straight line from $y$ to $x$:
    \begin{subequations}
        \begin{align}
        &\vert f(x)-f(y)\vert\\
        &=\Big\vert\int_y^x\nabla f(r) dr\Big\vert\\
        &=\Big\vert\int_0^1\langle \nabla f(xt+y(1-t)),x-y\rangle dt\Big\vert\\
        &\le\Big\vert\int_0^1\Vert \nabla f(xt+y(1-t))\Vert\cdot\Vert x-y\Vert dt\Big\vert\\
        &\le L_{f} \Big\vert\int_0^1\Vert x-y\Vert dt\Big\vert\\
        &=L_{f}\Vert x-y\Vert.
        \end{align}
    \end{subequations}
    The theorem follows.
\end{proof}

\noindent\textbf{Theorem 5.}
\textit{
Let $f_K:\mathbb{R}^n\rightarrow\mathbb{R}$ be a layer-wise 1-Lipschitz constrained $K$-layer network. The Lipschitz constant of the first $k$-layer network $L_{f_k}$ is upper-bounded by $L_{f_{k-1}}$, \ie,}
\begin{equation}
    L_{f_k}\le L_{f_{k-1}},\forall k\in \{2\cdots K\}.
\end{equation}
\begin{proof}
    Since all the layers including activation functions are all 1-Lipschitz constrained, \ie,
    \begin{equation}
        \begin{aligned}
            \Vert \mathbf{W}_k\cdot x-\mathbf{W}_k\cdot y\Vert&\le\Vert x-y\Vert,\forall x,y\in\mathbb{R}^{d_{k-1}}\\
            L_{\phi_k}&=1.
        \end{aligned}
        \label{lipschitz_layer_appendix}
    \end{equation}
    We can infer the upper bound of feature distance at layer $k$ by Eq.\eqref{lipschitz_layer_appendix}:
    \begin{equation}
        \begin{aligned}
            &\Vert f_k(x)-f_k(y) \Vert \\
            &=\Vert\phi_k(\mathbf{W}_k\cdot f_{k-1}(x)+\mathbf{b}_k)-\phi_k(\mathbf{W}_k\cdot f_{k-1}(y)+\mathbf{b}_k)\Vert \\
            &\le L_{\phi_k}\Vert(\mathbf{W}_k\cdot f_{k-1}(x)+\mathbf{b}_k)-(\mathbf{W}_k\cdot f_{k-1}(y)+\mathbf{b}_k)\Vert \\
            &\le L_{\phi_k}L_k\Vert f_{k-1}(x)-f_{k-1}(y)\Vert \\
            &= \Vert f_{k-1}(x)-f_{k-1}(y)\Vert. \\
        \end{aligned}
    \end{equation}
    This result implies 
    \begin{equation}
        \frac{\Vert f_k(x)-f_k(y)\Vert}{\Vert x-y\Vert}\le\frac{\Vert f_{k-1}(x)-f_{k-1}(y)\Vert}{\Vert x-y\Vert}, \forall x,y \in\mathbb{R}^n.
    \end{equation}
    The theorem follows.
\end{proof}

\noindent\textbf{Theorem 6.}
\textit{Let $f:\mathbb{R}^n\rightarrow\mathbb{R}$ be a continuous function which is modeled by a neural network, and all the activation functions of network $f$ are piecewise linear. The normalized function $\hat{f}(x)=f(x)/\big(\Vert\nabla_x f(x)\Vert+\vert f(x)\vert\big)$ is gradient norm bounded, \ie,}
\begin{equation}
    \Vert\nabla_x\hat{f}(x)\Vert=\Bigg\Vert\frac{\Vert\nabla f\Vert}{\Vert\nabla f\Vert+\vert f\vert}\Bigg\Vert^2\le 1.
\end{equation}
\begin{proof}
    For simplicity, function arguments are ignored here. By definition, the gradient norm of $\hat{f}(x)$ is:
    \begin{subequations}
        \begin{align}
            \Vert\nabla\hat{f}\Vert
            &=\Bigg\Vert\nabla\bigg(\frac{f}{\Vert\nabla f\Vert+\vert f\vert}\bigg)\Bigg\Vert \\
            &=\Bigg\Vert\frac{\nabla f\big(\Vert\nabla f\Vert+\vert f\vert\big)-f\nabla\big(\Vert\nabla f\Vert+\vert f\vert\big)}{\big(\Vert\nabla f\Vert+\vert f\vert\big)^2}\Bigg\Vert.
            \label{th5:pfeq2}
        \end{align}
    \end{subequations}
    By simple chain rule, we know that:
    \begin{subequations}
        \begin{align}
            \nabla\Vert\nabla f\Vert&=\nabla^2 f\frac{\nabla f}{\Vert\nabla f\Vert}, \\
            \nabla\vert f\vert&=\nabla f\frac{f}{\vert f\vert}.
        \end{align}
    \end{subequations}
    Since the network $f$ contains only piecewise linear activation functions, the Hessian matrix $\nabla^2 f$ is a zero matrix. The Eq.\eqref{th5:pfeq2} can be simplified:
    \begin{equation}
        \label{normalized_gradient_norm_appendix}
        \begin{aligned}
            \Vert\nabla\hat{f}\Vert
            =\Bigg\Vert\frac{\Vert\nabla f\Vert^2}{\big(\Vert\nabla f\Vert+\vert f\vert\big)^2}\Bigg\Vert
            =\Bigg\Vert\frac{\Vert\nabla f\Vert}{\Vert\nabla f\Vert+\vert f\vert}\Bigg\Vert^2\le 1.
        \end{aligned}
    \end{equation}
    The theorem follows.
\end{proof}
\section{Supplemental Experiments}

\noindent Please note that the source codes are archived for the verification in supplementary materials.

\subsection{Supplemental Ablation Study}
\begin{figure}[htbp]
    \centering
    \includegraphics[scale=0.20]{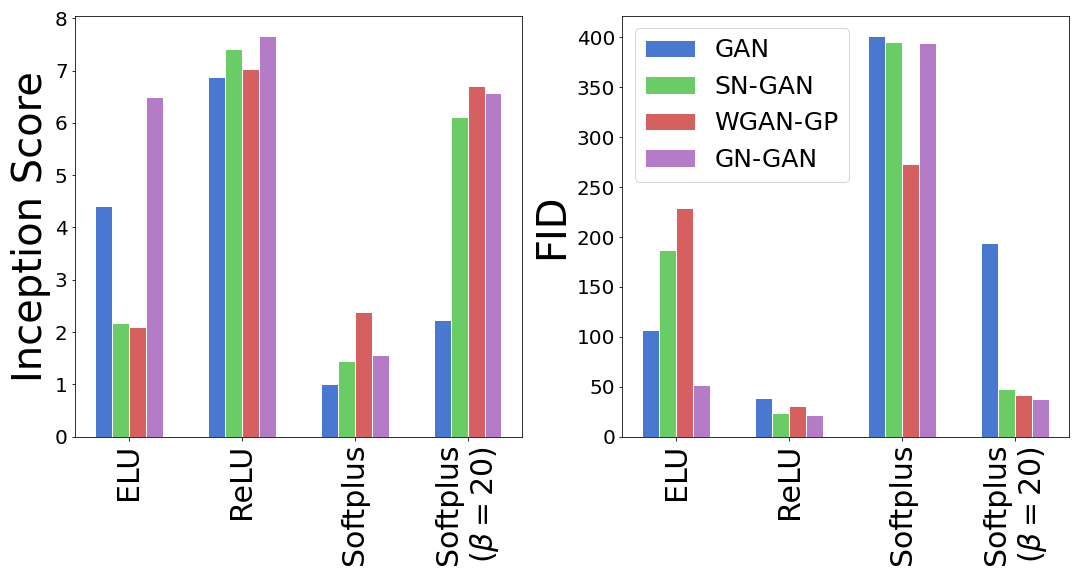}
    \caption{Comparison of activation functions including ELU~\cite{clevert2015fast}, ReLU~\cite{nair2010rectified} and Softplus~\cite{glorot2011deep} on CIFAR-10.}
    \label{ablation_activation_cifar10_appendix}
\end{figure}

\begin{figure}[htbp]
    \centering
    \includegraphics[scale=0.20]{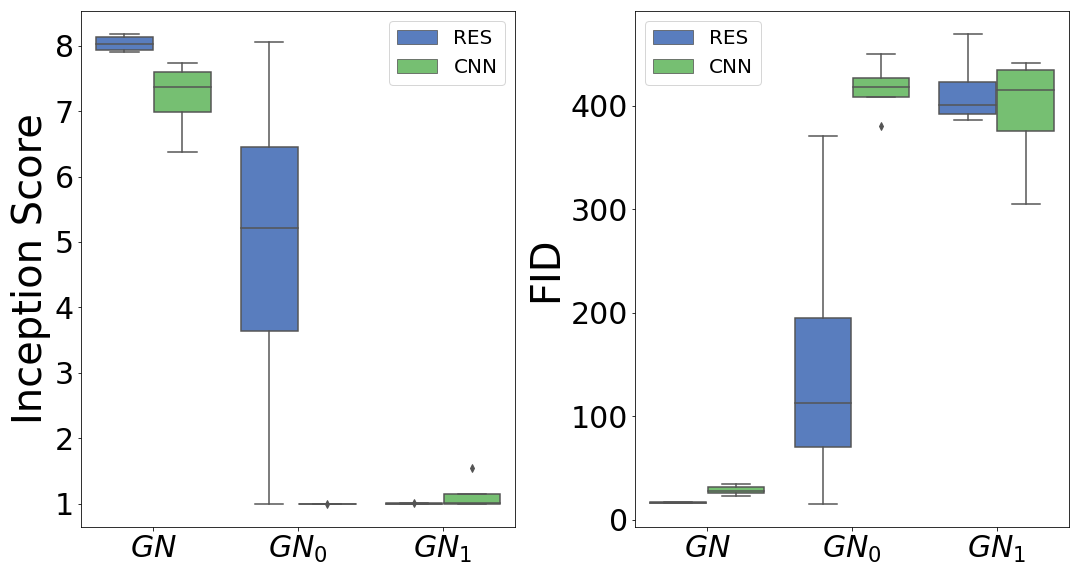}
    \caption{Comparison of variants of gradient normalization on CIFAR-10. The experiments include $\zeta(x)=\vert f(x)\vert$ ($GN$), $\zeta(x)=1$ ($GN_1$) and $\zeta(x)=0$ ($GN_0$).}
    \label{ablation_formulation_cifar10_appendix}
\end{figure}

Figure~\ref{ablation_activation_cifar10_appendix} compares the effectiveness of different activation functions in terms of IS and FID on CIFAR-10 dataset. The results show that the ReLU activation function achieves best IS and FID for different approaches. Moreover, the ReLU activation function with the proposed GN outperforms other state-of-the-art normalization and regularization approaches. It is worth noting that the original Softplus activation function achieves low IS and high FID for different approaches. However, by setting $\beta$ to 20, the result can be significantly better since Softplus becomes similar to ReLU if $\beta$ increases. Moreover, Figure~\ref{ablation_formulation_cifar10_appendix} compares the effectiveness of different $\zeta(x)$ in terms of IS and FID on CIFAR-10 dataset. The results indicate that the variance of Inception Score and FID for $GN_0$ is large for different architectures and datasets. The proposed $GN$ outperforms the alternatives, which is consistent to the experiments on STL-10 dataset.

\subsection{Decision Boundary Visualization}
\begin{figure*}[t!]
    \begin{subfigure}{0.16\textwidth}
        \centering
        \includegraphics[scale=0.55]{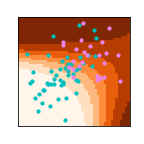}
        \caption{Theoretical}
        \label{value_surface_appendix:opt}
    \end{subfigure}%
    \begin{subfigure}{0.16\textwidth}
        \centering
        \includegraphics[scale=0.55]{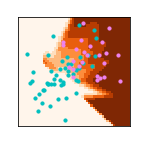}
        \caption{Vanilla GAN}
        \label{value_surface_appendix:gan}
    \end{subfigure}%
    \begin{subfigure}{0.16\textwidth}
        \centering
        \includegraphics[scale=0.55]{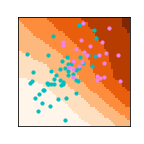}
        \caption{0-GP}
        \label{value_surface_appendix:gp0}
    \end{subfigure}%
    \begin{subfigure}{0.16\textwidth}
        \centering
        \includegraphics[scale=0.55]{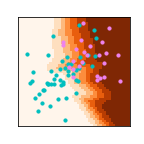}
        \caption{1-GP}
        \label{value_surface_appendix:gp1}
    \end{subfigure}%
    \begin{subfigure}{0.16\textwidth}
        \centering
        \includegraphics[scale=0.55]{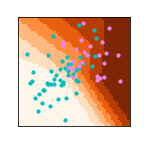}
        \caption{SN}
        \label{value_surface_appendix:sn}
    \end{subfigure}%
    \begin{subfigure}{0.2\textwidth}
        \centering
        \includegraphics[scale=0.55]{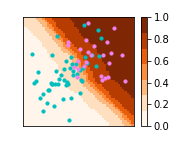}
        \caption{GN}
        \label{value_surface_appendix:gn}
    \end{subfigure}%
    \caption{The theoretical and empirical value surfaces of discriminators which are parameterized by a 2-layer MLP with hidden size 512. Real samples are drawn from a 2D multivariate Gaussian and fixed for all discriminators, while the fake samples are sampled from the other 2D multivariate Gaussian infinitely. (\subref{value_surface_appendix:opt}) The theoretically optimal discriminator $D^*(x)=p_r(x)/(p_r(x)+p_g(x))$. (\subref{value_surface_appendix:gn}) Our gradient normalization.}
    \label{value_surface_appendix}
\end{figure*}

We conduct an experiment similar to~\cite{thanh2019improving} for the visualization. The value surfaces of binary classification tasks are demonstrated in Figure \ref{value_surface_appendix}. The results demonstrate that the value surface of vanilla GAN (Figure \ref{value_surface_appendix:gan}) contains steep cliffs near to the decision boundary, which causes gradient explosion when the synthetic samples are located in this area. With the regularization or normalization applied to discriminator, the value surface becomes smooth in varying levels as shown in Figures~\ref{value_surface_appendix}(\subref{value_surface_appendix:gp0})-(\subref{value_surface_appendix:gn}).

\subsection{Training Speed}
Table~\ref{training_speed_appendix} shows the training speed of different approaches with ResNet as the backbone network on CIFAR-10 dataset. All the training processes are performed on NVIDIA RTX 2080Ti five times, and we report the average results in terms of update iterations per second. The results show that different approaches require additional computation as compared to the Vanilla GAN. It is worth noting that although the training speed of the proposed GN is only compatible with 1-GP, the proposed GN outperforms the other approaches in terms of IS and FID. In other words, even with more computation, other approaches can not improve their results. On the other hand, the training process is offline, while the inference speed is the same for different approaches.
\renewcommand{\arraystretch}{1.15}
\begin{table}[hbt!]
    \centering
    \begin{tabular}{c|cc}
        \hline
        Method   & Generator (it/s) & Discriminator (it/s) \\
        \hline
        Vanilla  & 6.91 & 15.73 \\
        SN       & 6.52 & 14.41 \\
        1-GP     & 4.70 & 7.66 \\
        GN       & 3.68 & 6.48 \\
        \hline
    \end{tabular}
    \caption{Training speed of generator and discriminator.}
    \label{training_speed_appendix}
\end{table}

\subsection{Loss Function Comparison}
We further investigate the performance of the proposed GN with different loss functions. Notably, the Gradient Normalization makes the outputs of discriminators saturate in range $[-1, 1]$, and thus the sigmoid at the end of discriminator can be eliminated when the non-saturating loss is used. Moreover, the hinge loss is equivalent to Wasserstein loss in the perspective of gradients when GN is used, \ie,
\begin{equation}
    \begin{aligned}
        \nabla\mathcal{L}_{hinge}=&\nabla\mathbb{E}_{x\sim p_g(x)}[\max(1+\hat{D}(x),0)]+\\
        &\nabla\mathbb{E}_{x\sim p_r(x)}[\max(1-\hat{D}(x),0)] \\
        =&\nabla\mathbb{E}_{x\sim p_g(x)}[1+\hat{D}(x)]+\\
        &\nabla\mathbb{E}_{x\sim p_r(x)}[1-\hat{D}(x)] \\
        =&\nabla\mathbb{E}_{x\sim p_g(x)}[\hat{D}(x)]-\nabla\mathbb{E}_{x\sim p_r(x)}[\hat{D}(x)] \\
        =&\nabla\mathcal{L}_{wasserstein}
    \end{aligned}
\end{equation}

Table~\ref{loss_comparison_appendix} shows the evaluation results of different loss functions on CIFAR-10 in terms of Inception score and FID. Both ResNet and CNN architectures are reported. Since the Wasserstein loss is equivalent to hinge loss, the Wasserstein loss is not listed. The performance of GN-GANs is consistent with different loss functions.

\renewcommand{\arraystretch}{1.15}
\begin{table}[t!]
    \centering
    \begin{tabular}{lccc}
        \hline
        \textbf{Loss Function} &  IS$\uparrow$ & FID(train)$\downarrow$ & FID(test)$\downarrow$ \\
        \hline
        \multicolumn{4}{l}{\textit{\textbf{Standard CNN}}} \\
        Hinge & 7.67$\pm$.14 & 18.20$\pm$.12 & 22.24$\pm$.88 \\
        NS & 7.78$\pm$.11 & 18.17$\pm$.61 & 22.36$\pm$.59 \\
        NS$_{-sigmoid}$ & 7.69$\pm$.16 & 18.93$\pm$.87 & 23.19$\pm$.86 \\
        \hline
        \multicolumn{4}{l}{\textit{\textbf{ResNet}}} \\
        Hinge & 8.49$\pm$.11 & 11.13$\pm$.18 & 15.33$\pm$.16 \\
        NS & 8.49$\pm$.11 & 10.97$\pm$.22 & 15.15$\pm$.29 \\
        NS$_{-sigmoid}$ & 8.49$\pm$.09 & 11.01$\pm$.26 & 15.14$\pm$.32 \\
        \hline
    \end{tabular}
    \caption{Loss function comparison of GN-GANs on CIFAR-10. Note that the non-saturating loss without sigmoid at the last layer is denoted by NS$_{-sigmoid}$.}
    \label{loss_comparison_appendix}
\end{table}
\section{Evaluation Details}
\label{appendix_evaluation_details}

\noindent\textbf{Inception Score.}
For the Inception Score (IS), we divide 50k generated images into 10 partitions and calculate the average and the standard deviation of Inception Score over each partition. The final results are the average scores of different training sessions.

\noindent\textbf{Frechet Inception Distance.}
The configurations of FID are described as follow. For the CIFAR-10 dataset, we use 50k generated samples vs.\ 50k training images and 10k generated samples vs.\ 10k test images. For the STL-10 dataset, we use 50k generated samples vs.\ 100k unlabeled images and 10k generated samples vs.\ 100k unlabeled images. For the CelebA-HQ, we use 30k generated samples vs.\ 30k training images. For the LSUN Church Outdoor, we use 50k generated samples vs.\ 126k training images. In the training process, models are trained on CIFAR-10 training set, STL-10 unlabeled images, CelebA-HQ training set and LSUN Church Outdoor training set.

\section{Experimental Details}
\label{appendix_experiments_details}

\begin{figure*}[p]
    \centering
    \begin{subfigure}{0.49\textwidth}
        \centering
        \includegraphics[scale=0.65]{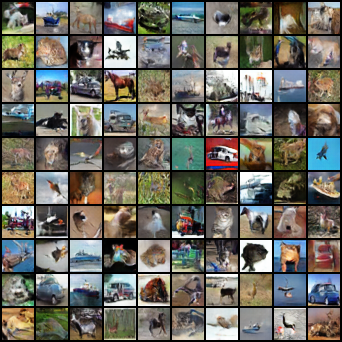}
        \caption{GN-GAN CNN}
    \end{subfigure}
    \begin{subfigure}{0.49\textwidth}
        \centering
        \includegraphics[scale=0.65]{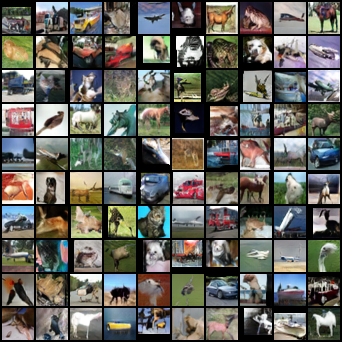}
        \caption{GN-GAN-CR CNN}
    \end{subfigure}
    
    \begin{subfigure}{0.49\textwidth}
        \centering
        \includegraphics[scale=0.65]{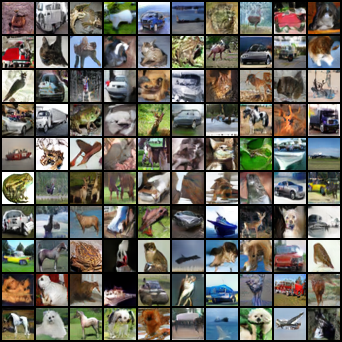}
        \caption{GN-GAN ResNet}
    \end{subfigure}
    \begin{subfigure}{0.49\textwidth}
        \centering
        \includegraphics[scale=0.65]{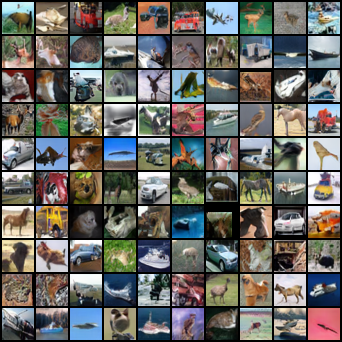}
        \caption{GN-GAN-CR ResNet}
    \end{subfigure}
    \caption{Unconditional image generation on CIFAR-10.}
    \label{unconditional_cifar_10_figure_appendix}
\end{figure*}

\begin{figure*}[p]
    \centering
    \begin{subfigure}{0.49\textwidth}
        \centering
        \includegraphics[scale=0.45]{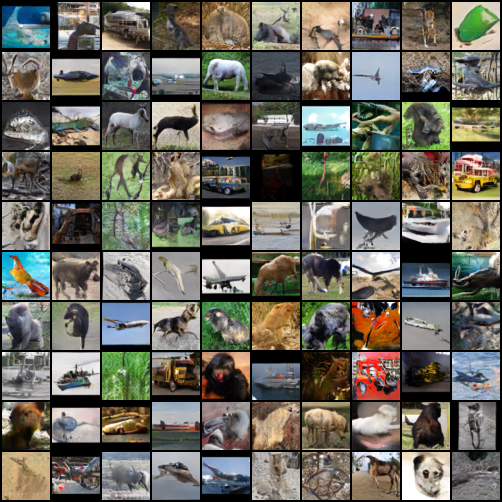}
        \caption{GN-GAN CNN}
    \end{subfigure}
    \begin{subfigure}{0.49\textwidth}
        \centering
        \includegraphics[scale=0.45]{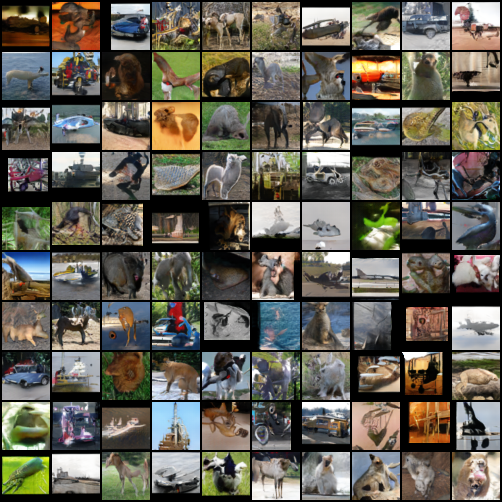}
        \caption{GN-GAN-CR CNN}
    \end{subfigure}
    
    \begin{subfigure}{0.49\textwidth}
        \centering
        \includegraphics[scale=0.45]{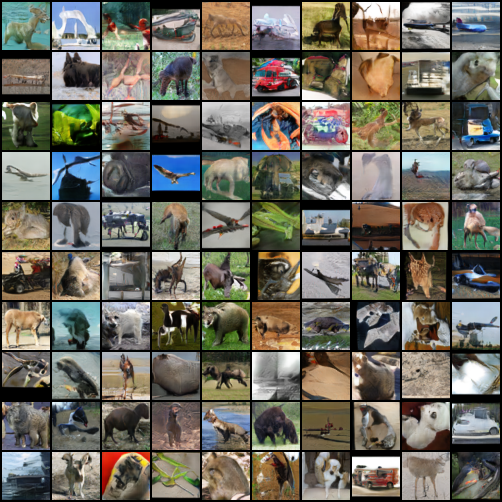}
        \caption{GN-GAN ResNet}
    \end{subfigure}
    \begin{subfigure}{0.49\textwidth}
        \centering
        \includegraphics[scale=0.45]{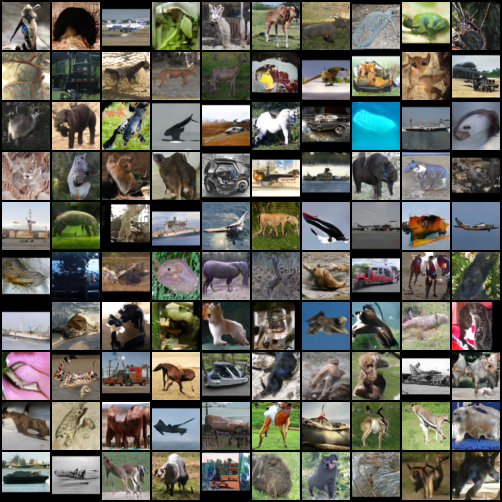}
        \caption{GN-GAN-CR ResNet}
    \end{subfigure}
    \caption{Unconditional image generation on STL-10.}
    \label{unconditional_stl_10_figure_appendix}
\end{figure*}

\noindent\textbf{Unconditional Image Generation on CIFAR-10 and STL-10.}
For the fair comparison, we use the ResNet architecture as well as the Standard CNN used in~\cite{Miyato2018SpectralNF}. The last layer of ResNet, \ie, global sum pooling, is replaced by the global average pooling. Moreover, all the weights of fully-connected layers and CNN layers are initialized by Kaiming Normal Initialization~\cite{he2015delving}, and the biases are initialized to zero. We use Adam~\cite{Kingma2015AdamAM} as the optimizer with parameters $\alpha_G=2\times10^{-4}$, $\alpha_D=4\times10^{-4}$, $\beta_1=0$, $\beta_2=0.9$ and batch size $M=64$. The learning rate linearly decays to $0$ through the training. The generator is updated once for every 5 discriminator update steps. All the training processes are stopped after the generator update 200k steps. For the data augmentation, the random horizontal flipping is applied for every method (including our method and re-implementation). The augmentation setting in Table~\ref{cr_pipeline_table_appendix} is used for Consistency Regularization~\cite{Zhang2020ConsistencyRF}. For more qualitative results, please refer to Figures~\ref{unconditional_cifar_10_figure_appendix} and ~\ref{unconditional_stl_10_figure_appendix}.

\renewcommand{\arraystretch}{1.15}
\begin{table}[hbt!]
    \centering
    \begin{tabular}{|c|l|}
        \hline
         1. & RandomHorizontalFlipping(p=$0.5$)\\
        \hline
         2. & RandomPixelShifting(pixel=$0.2\times$ImageSize) \\
        \hline
    \end{tabular}
    \caption{Augmentation for consistency regularization on CIFAR-10 and STL-10.}
    \label{cr_pipeline_table_appendix}
\end{table}

\noindent\textbf{Conditional Image Generation on CIFAR-10.} To show the results of conditional image generation on CIFAR-10 dataset, we compare the results of BigGAN~\cite{Brock2019LargeSG}, BigGAN with the Consistency Regularization (CR), BigGAN with the proposed GN, and BigGAN with the proposed GN and CR. Here, the discriminator in the conditional GAN is considered as a conditional function, \ie, $D_y(x)$, instead of the multi-variable function, \ie, D($x$, $y$). Therefore, the Gradient Normalization can be formulated as follows:
\begin{equation}
    \hat{D}_y(x)=\frac{D_y(x)}{\Vert\nabla_x D_y(x)\Vert+\Vert D_y(x)\Vert},
\end{equation}
where $D_y(x)$ is the discriminator conditional on $y$. Similarly, by Theorem 5, $\hat{D}_y(x)$ is a gradient norm bounded network with respect to $x$.

\begin{figure*}[p]
    \centering
    \includegraphics[scale=0.7]{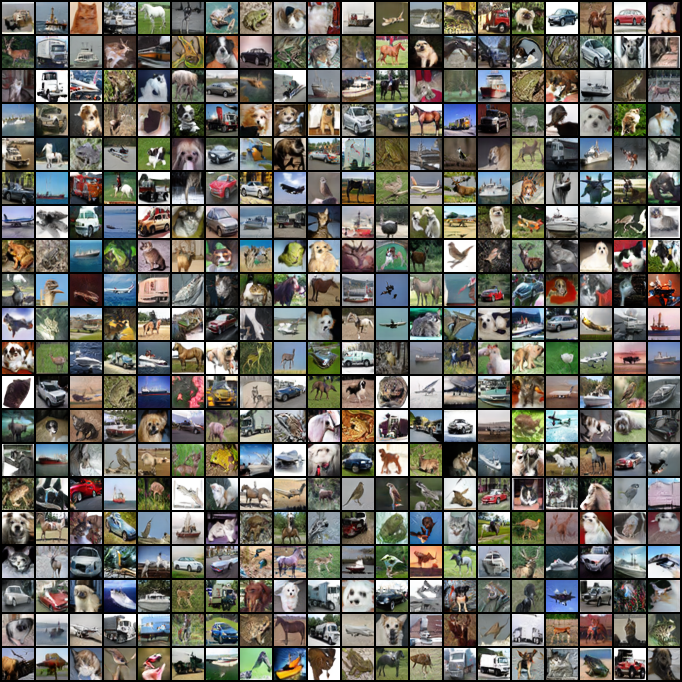}
    \caption{Conditional image generation on CIFAR-10.}
    \label{contidional_cifar10_figure_appendix}
\end{figure*}

Moreover, we take the official implementation of BigGAN~\cite{Brock2019LargeSG} for the reference. We use Adam as the optimizer with parameters $\alpha_G=1\times10^{-4}$, $\alpha_D=2\times10^{-4}$, $\beta_1=0$, $\beta_2=0.999$ and the batch size as $50$. The generator is updated once for every 4 discriminator update steps. All the training processes are stopped after the generator updates $125k$ steps. The real images are augmented by the random horizontal flipping. Following the previous setting~\cite{karras2017progressive,Mescheder2018WhichTM}, we employ the moving averages on generator weights with a decay of $0.9999$. The pipeline for CR is shown in Table~\ref{cr_pipeline_table_appendix}. Table~\ref{conditional_image_generation_table_appendix} shows the performance of different approaches in terms of IS, FID (train) and FID (test). The results indicate that BigGAN with the proposed GN is better than BigGAN with CR, while BigGAN with both GN and CR achieves the best performance. For more qualitative results, please refer to Figure~\ref{contidional_cifar10_figure_appendix}.

\setlength{\tabcolsep}{4pt}
\renewcommand{\arraystretch}{1.15}
\begin{table}[hbt!]
    \centering
    \caption{Inception Score(IS) and FID of conditional image generation on CIFAR-10.}
    \begin{tabular}{lccc}
        \hline
        Method & IS$\uparrow$ & FID(train)$\downarrow$ & FID(test)$\downarrow$ \\
        \hline
        BigGAN~\cite{Brock2019LargeSG} & 9.22 & - & 14.73\\
        BigGAN-CR~\cite{Zhang2020ConsistencyRF} & - & - & 11.48 \\
        GN-BigGAN & 9.22$\pm$.13 & 5.87$\pm$.15 & 10.05$\pm$.23 \\
        GN-BigGAN-CR & \textbf{9.35$\pm$.14} & \textbf{4.86$\pm$.07} & \textbf{8.92$\pm$.15} \\
        \hline
    \end{tabular}
    \label{conditional_image_generation_table_appendix}
\end{table}

\begin{figure*}[p]
    \centering
    \includegraphics[scale=0.45]{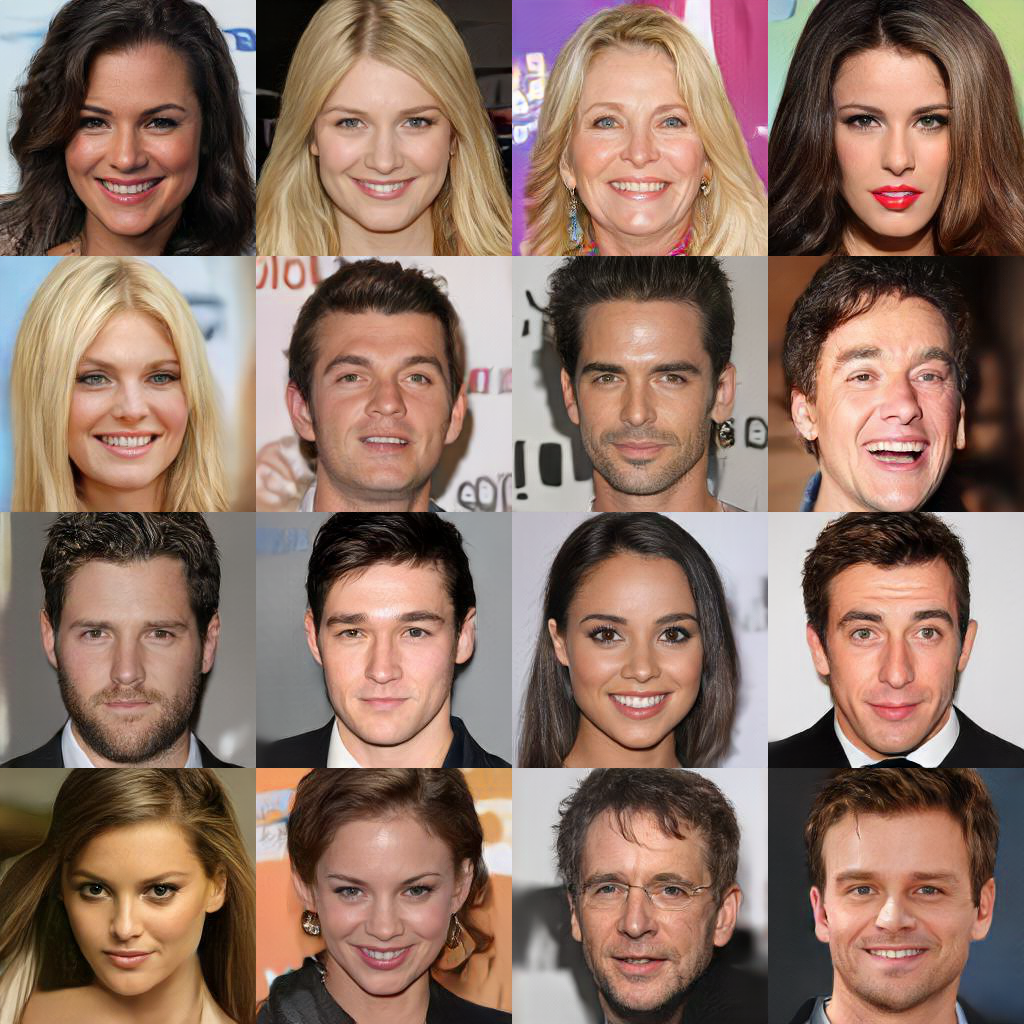}
    \caption{Unconditional image generation on CelebA-HQ $256\times256$.}
    \label{celebhq_7x7_figure_appendix}
\end{figure*}

\begin{figure*}[p]
    \centering
    \includegraphics[scale=0.45]{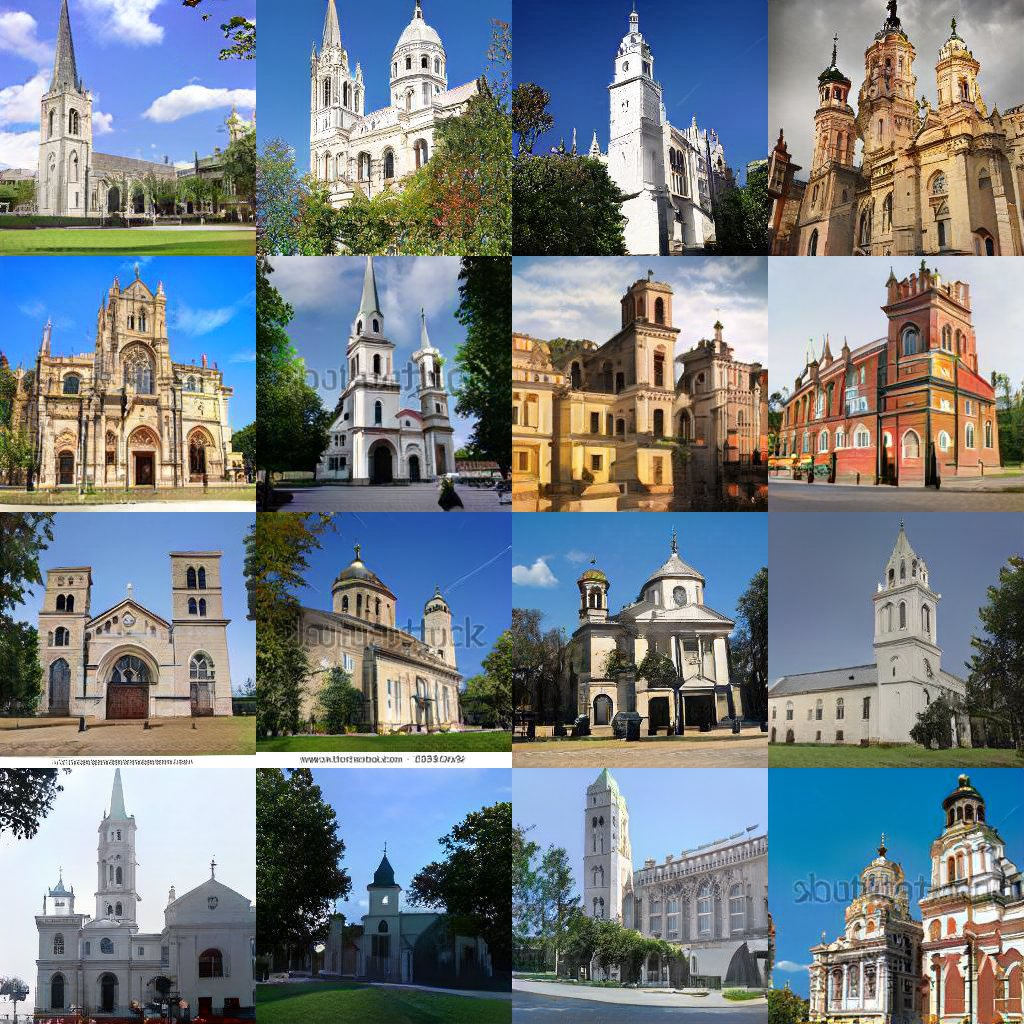}
    \caption{Unconditional image generation on LSUN Church Outdoor $256\times256$.}
    \label{church_7x7_figure_appendix}
\end{figure*}

\noindent\textbf{Unconditional Image Generation on CelebA-HQ and LSUN Church Outdoor.} We further evaluate the proposed Gradient Normalization on two high-resolution image datasets, i.e., CelebA-HQ and LSUN Church Outdoor. For the augmentation, the random horizontal flipping is 
adopted for both datasets.
We use the architecture proposed by SN-GAN~\cite{Miyato2018SpectralNF} for generating $256\times256$ images. We use Adam again as the optimizer with parameters $\alpha_G=2\times10^{-4}$, $\alpha_D=2\times10^{-4}$, $\beta_1=0$, $\beta_2=0.9$ and batch size as $64$. The generator is updated once for every 5 discriminator update steps. All the training processes are stopped after the generator update 100k steps. We employ the moving averages on generator weights with a decay of 0.9999. The Inception Score and FID are shown in Table~\ref{large_scale_unconditional_image_generation_table_appendix}. It is worth noting that the performance can be further improved with a better architecture. For more qualitative results, please refer to Figures~\ref{celebhq_7x7_figure_appendix} and~\ref{church_7x7_figure_appendix}.

\renewcommand{\arraystretch}{1.15}
\begin{table}[hbt!]
    \centering
    \caption{Inception Score and FID of unconditional image generation on CelebA-HQ and LSUN Church Outdoor. $\dagger$ represents that we provide SN-GAN implementation as the baseline.}
    \begin{tabular}{lcc}
        \hline
        Dataset         & GN-GAN         & SN-GAN            \\
        \hline
        CelebA-HQ 128   & \textbf{14.78} & 25.95 (from [30]) \\
        CelebA-HQ 256   & \textbf{7.67}  & 14.45$^\dagger$   \\
        LSUN Church 256 & \textbf{5.41}  & 8.44$^\dagger$    \\
        \hline
    \end{tabular}
    \label{large_scale_unconditional_image_generation_table_appendix}
\end{table}

\begin{figure*}[p]
    \centering
    \includegraphics[scale=0.225]{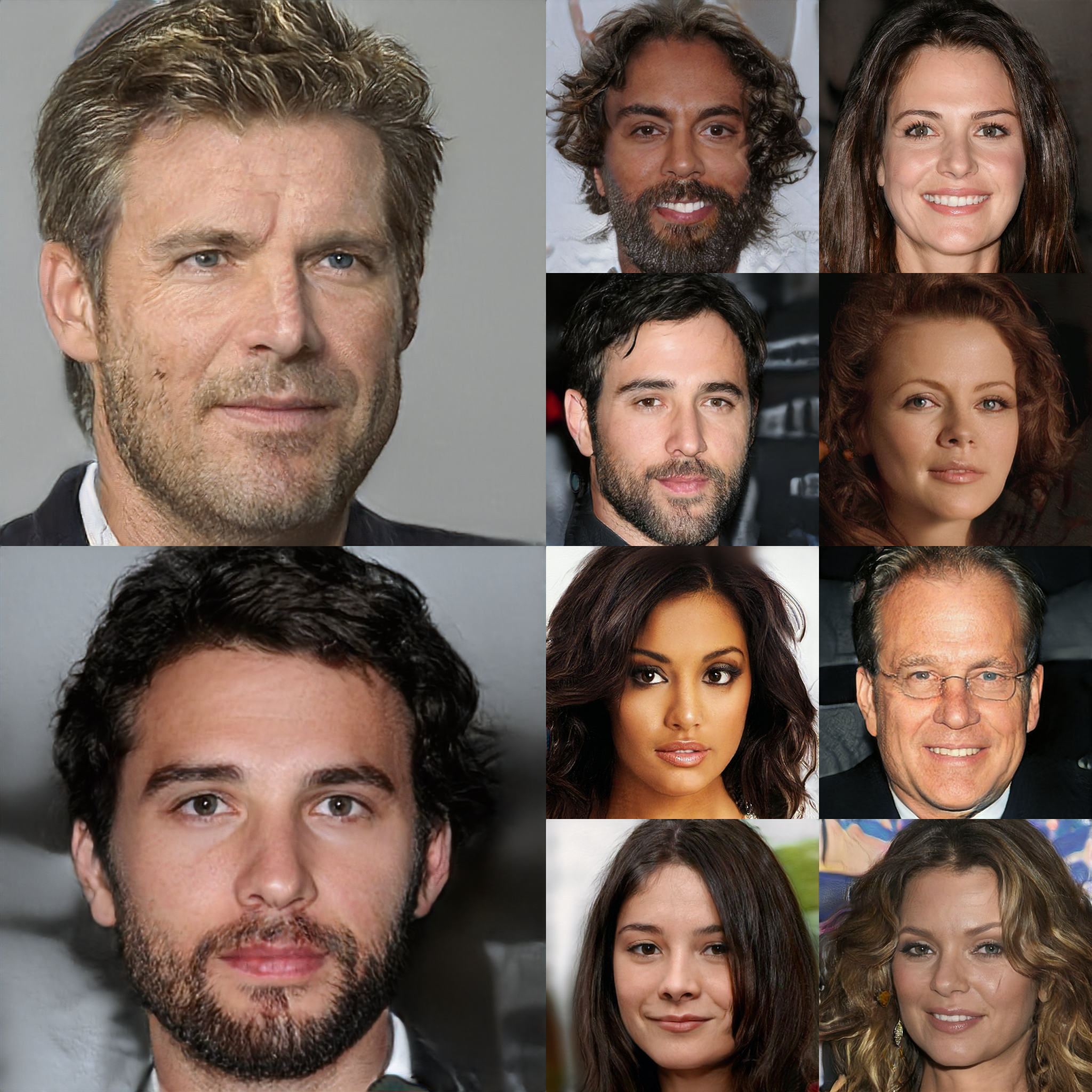}
    \caption{GN-StyleGAN on CelebA-HQ $1024\times1024$.}
    \label{stylegan_grid_figure_appendix}
\end{figure*}

\begin{figure*}[p]
    \centering
    \includegraphics[scale=0.4]{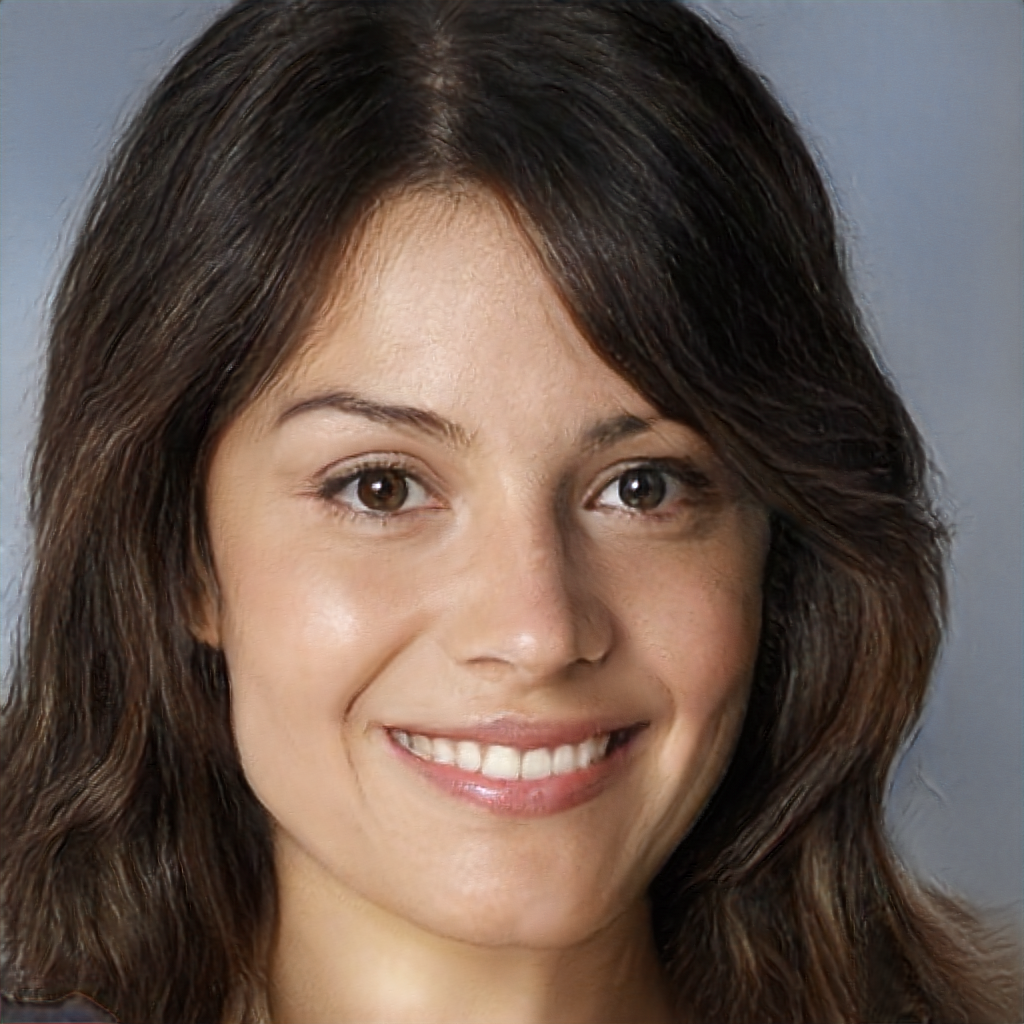}
    \caption{GN-StyleGAN on CelebA-HQ $1024\times1024$.}
    \label{stylegan_large1_figure_appendix}
\end{figure*}

\begin{figure*}[p]
    \centering
    \includegraphics[scale=0.4]{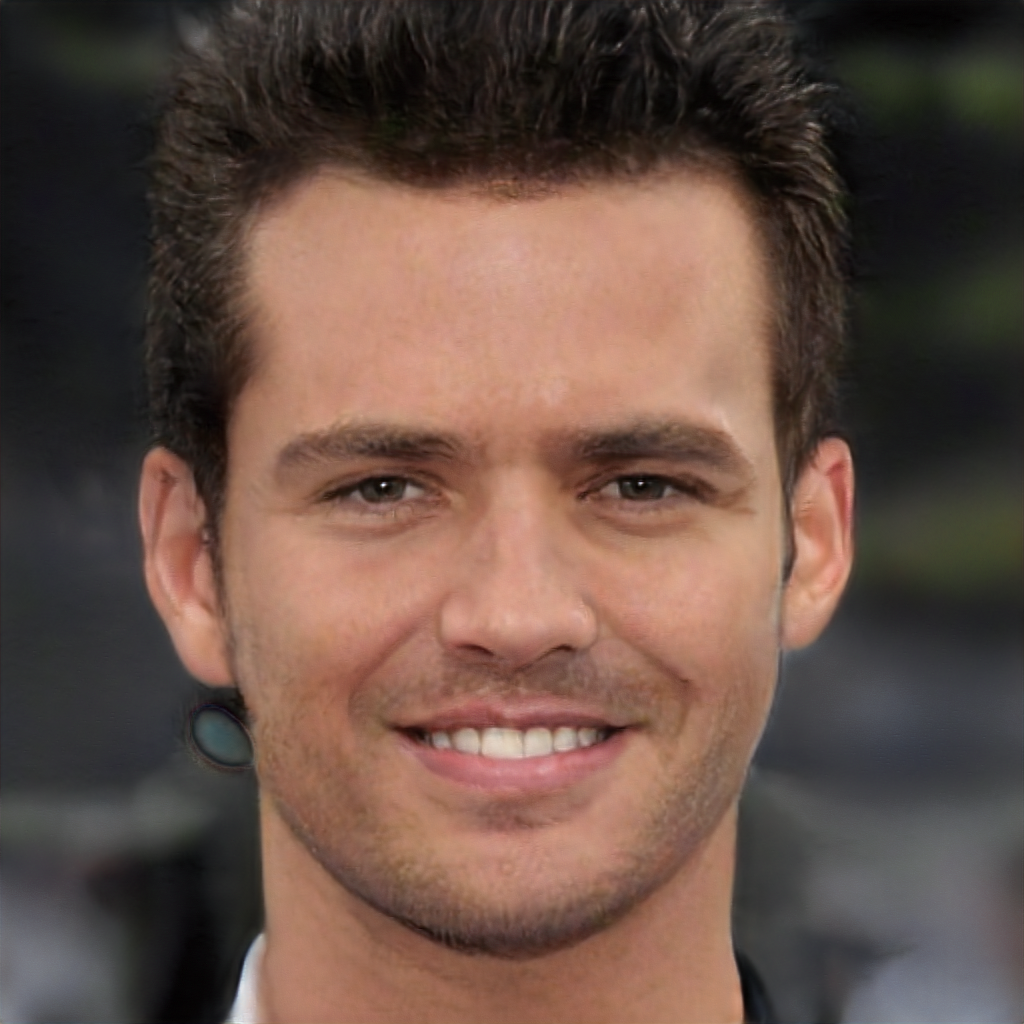}
    \caption{GN-StyleGAN on CelebA-HQ $1024\times1024$.}
    \label{stylegan_large2_figure_appendix}
\end{figure*}

\begin{figure*}[p]
    \centering
    \includegraphics[scale=0.4]{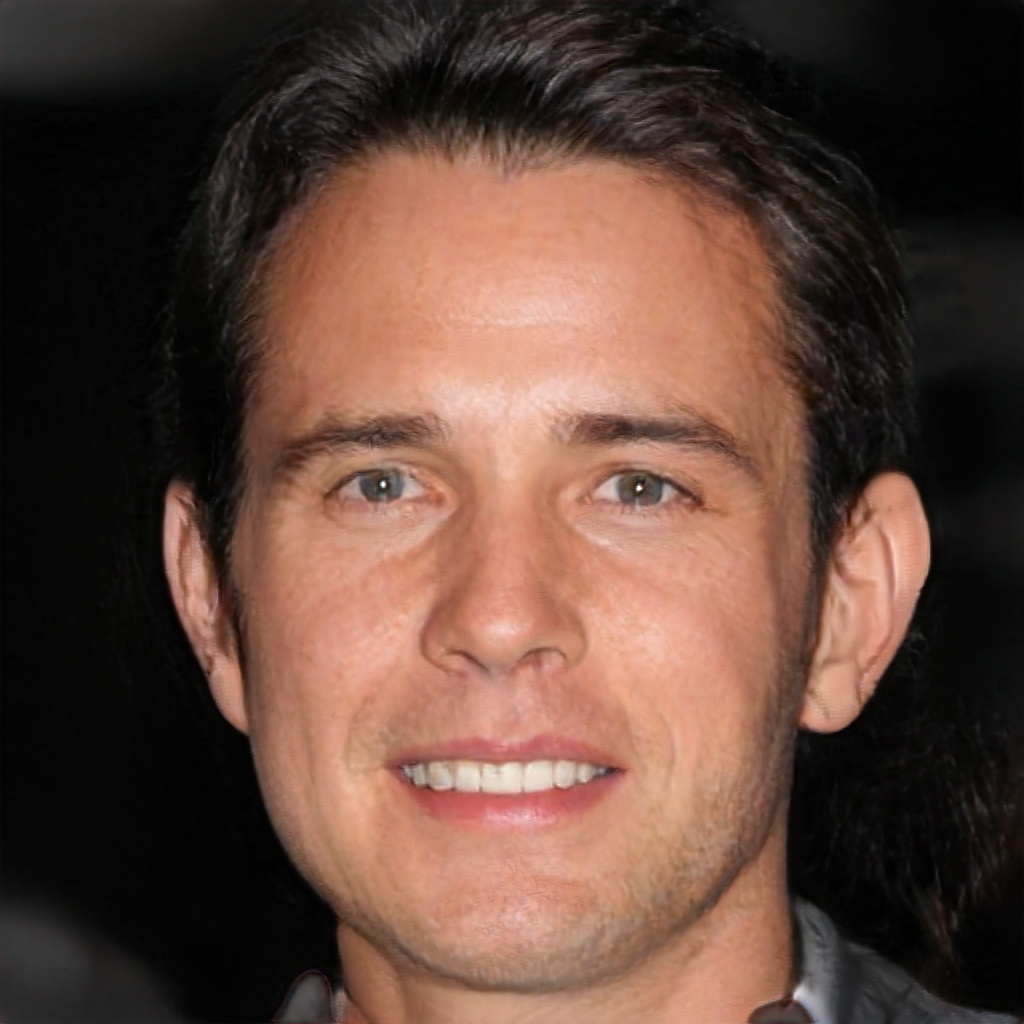}
    \caption{GN-StyleGAN on CelebA-HQ $1024\times1024$.}
    \label{stylegan_large3_figure_appendix}
\end{figure*}

\noindent\textbf{Experiments on Progressive Growing Architecture.} We further test the StyleGAN~\cite{Karras_2019_CVPR} with the proposed Gradient Normalization on CelebA-HQ $1024\times1024$. Note that the R1 regularization and Gradient Penalty are replaced with GN in our experiment. We use hinge loss as the objective function and Adam as the optimizer. The learning rates $\alpha_G$ and  $\alpha_D$ are both set to $0.001$ for resolutions of $8^2$, $16^2$, $32^2$ and $64^2$, and $0.0015$ otherwise. For the other settings, we use the same parameters as StyleGAN. The FID of GN-StyleGAN is $8.65$ which is calculated by 50k generated images vs.\ 30k training images. The generated samples are shown in Figures~\ref{stylegan_grid_figure_appendix}-\ref{stylegan_large3_figure_appendix}.

\clearpage

{\small
\bibliographystyle{ieee_fullname}
\bibliography{egbib}
}

\end{document}